\documentclass[runningheads]{llncs}

\usepackage{eccv}

\usepackage{eccvabbrv}

\usepackage{graphicx}
\usepackage{booktabs}
\usepackage{multirow}
\usepackage{adjustbox}
\usepackage{amsmath,amssymb}
\usepackage{hyperref}
\usepackage[accsupp]{axessibility}
\usepackage{orcidlink}
\usepackage{placeins}
\usepackage[expansion=false]{microtype} 

\begin{document}

\title{Integrated Forward--Inverse Network \texorpdfstring{\\}{ }for Lensless Image Reconstruction}

\titlerunning{Integrated Forward--Inverse Network}


\author{
Donggeon Bae\inst{1}\orcidlink{0009-0003-4913-0009}\and
Jaewoo Jung\inst{2,3}\orcidlink{0000-0003-4340-6193}\and
Yong Guk Kang\inst{2}\orcidlink{0000-0002-5572-7544}\and
Kyung Chul Lee\inst{4}\orcidlink{0000-0002-5533-3078}\and
Taeyoung Kim\inst{3}\orcidlink{0000-0003-2774-722X}\and
Jongho Kim\inst{1}\orcidlink{0009-0001-2380-7945}\and
Sangjun Byun\inst{1}\orcidlink{0009-0005-7098-4337}\and
Joonsik Park\inst{3}\orcidlink{0009-0004-8844-4868}\and
Seung Ah Lee\inst{1,2}\orcidlink{0000-0001-5173-1565}\thanks{Corresponding author.}
}
\authorrunning{D.~Bae et al.}
\institute{
Department of Mechanical Engineering, Seoul National University, Republic of Korea \and
School of Mechanical and Aerospace Engineering/SNU-IAMD, Seoul National University, Republic of Korea \and
Department of Electrical and Electronic Engineering, Yonsei University, Republic of Korea \and
Department of Biomedical Engineering, University of Michigan, Ann Arbor, MI, USA\\
\email{\{donggeonbae, seungahlee\}@snu.ac.kr}
}

\maketitle
\begin{abstract} Lensless imaging enables compact and versatile computational cameras by replacing bulky optics with thin coded elements. However, reconstruction from the resulting measurements is challenging: large-footprint point-spread functions (PSFs) produce highly multiplexed observations, making inversion severely ill-conditioned and sensitive to calibration errors and model mismatch. While deep learning approaches, including hybrid models that incorporate physics priors, have shown promise, explicitly maintaining data fidelity throughout the network hierarchy remains difficult. Here, we propose the Integrated Forward--Inverse Network (IFIN), a physics-guided architecture that interleaves differentiable forward projections with learnable inverse updates at every scale, enabling complementary cues to be exploited jointly in the measurement and image domains. This bidirectional coupling supports progressive, physics-consistent refinement and permits system-constrained PSF kernel adaptation under model uncertainty. On challenging lensless benchmarks, including a newly introduced dataset, IFIN achieves state-of-the-art reconstruction quality. We further observe competitive performance on Gaussian deblurring and simulated inline holography reconstruction, suggesting that the same interleaving principle can extend beyond lensless cameras.

\end{abstract}

\section{Introduction}

Modern optical imaging systems, ranging from compact lensless cameras with coded apertures to advanced microscopes with engineered point-spread functions (PSFs), are increasingly designed with complex forward models. A growing class of such systems operates with PSFs that are intentionally or unavoidably broadened by designed optical coding. In this regime, each measurement mixes scene information over a wide spatial extent, making the inverse problem severely ill-conditioned. While such designs unlock diverse imaging capabilities that transcend the limits of conventional optics~\cite{sahoo2017single, satat2017object, antipa2017diffusercam, antipa2019video, baek2022lensless}, they also introduce substantial challenges for reconstruction. In practice, the effective PSFs can vary across both the field of view and channels, violating the stationarity assumptions underlying standard inverse pipelines~\cite{thiebaut2016spatially, yanny2022deep, cai2024phocolens}. Hardware imperfections and residual modulations further deteriorate this model mismatch, making accurate and robust reconstruction a central challenge as optical platforms continue to shrink and diversify.

A wide spectrum of approaches has been explored for image reconstruction in lensless imaging systems. Classical inverse mappings~\cite{wiener1964extrapolation} and model-based optimization methods~\cite{richardson1972bayesian, lucy1974iterative, boyd2011distributed} built on well-defined priors offer physically grounded results, but they are often computationally expensive, sensitive to calibration errors, and unreliable under model mismatch. With advances in deep learning, data-driven methods~\cite{bae2020lensless, pan2022image} have enabled end-to-end mappings from measurements to target scenes, yet they may not explicitly encode the underlying system physics, which can reduce accuracy and robustness under out-of-distribution conditions; such models may also produce hallucinations. In response, hybrid methods~\cite{monakhova2019learned, yanny2022deep, kingshott2022unrolled, li2023mwdns} that embed the physical forward model within a learning framework have emerged, improving efficiency and grounding predictions in the physical model while leveraging data-driven components to capture priors that are difficult to specify analytically.

However, many existing hybrid pipelines incorporate physics in a one-sided manner, either operating primarily in the measurement domain or refining only an already inverted estimate; once an image is reconstructed, the measurement cues become less accessible. Consequently, measurement information may collapse as it passes through the network, and intermediate estimates can become detached from the raw measurements. This becomes especially problematic in practical lensless reconstruction, where the effective forward model varies across the field of view and is only approximately calibrated. In this setting, whether inversion is applied after this collapse or refinement follows an inversion without measurement-domain feedback, mismatch-induced errors can manifest as spatially structured artifacts and persist through later stages.

A key opportunity in this setting is to leverage physics priors and learned priors jointly throughout reconstruction, rather than confining physical consistency checks to a single stage. In optical systems with broad PSFs and long-range mixing, preserving measurement-domain cues while refining image-domain representations provides complementary information for recovering fine details. To this end, we introduce \textbf{IFIN}, a bidirectional reconstruction framework that interleaves differentiable forward projections with learnable inverse updates within an encoder--decoder hierarchy, enabling stable, physics-consistent reconstruction under large-footprint PSFs and shift-variant degradations. IFIN further learns a shift-variant PSF field end-to-end, improving robustness to calibration mismatch and enabling blind recovery when PSFs are inaccurate or unavailable.

Compared to the prior state of the art, IFIN improves PSNR on the three lensless benchmarks by +1.63\,dB (DiffuserCam~\cite{monakhova2019learned}), +0.65\,dB (\textbf{WiderCam}, our newly introduced lensless benchmark), and +2.58\,dB (MultiWienerNet~\cite{yanny2022deep}). Beyond lensless imaging, IFIN remains competitive on simulated Gaussian deblurring and achieves strong gains on inline holography reconstruction, demonstrating that the proposed bidirectional forward--inverse integration generalizes beyond a single modality and can be applied to a broader class of inverse problems.

Our main contributions are as follows:
\begin{enumerate}
  \item We propose IFIN, a reconstruction framework that embeds \emph{bidirectional} forward--inverse guidance at \emph{every} encoder--decoder scale, repeatedly exchanging measurement- and image-domain cues for physics- and data-driven refinement under long-range mixing.
  \item Within this framework, IFIN jointly learns a shift-variant PSF field, shared by both operators across scales, for robustness to calibration mismatch and blind recovery.
  \item IFIN achieves state-of-the-art performance on three lensless benchmarks and further validates the proposed approach on simulated Gaussian deblurring and inline holography.
\end{enumerate}

\section{Related Work}
\label{sec:related_work}

\subsection{Lensless Imaging}
Lensless cameras replace conventional lenses with thin optical elements such as coded apertures~\cite{asif2016flatcam}, transmissive diffusers~\cite{antipa2017diffusercam}, and engineered phase masks~\cite{boominathan2020phlatcam, lee2023design}. As a result, diffuser- or mask-induced PSFs are large and highly structured, often encoding wide spatial neighborhoods onto the sensor, up to the entire scene. This encoding eliminates the need for bulky optics but necessitates computational reconstruction to recover interpretable images from the raw measurements.

Beyond simple image recovery, lensless systems support diverse modalities---including depth~\cite{antipa2017diffusercam, bagadthey2022flatnet3d}, hyperspectral~\cite{sahoo2017single, monakhova2020spectral}, polarization~\cite{baek2022lensless}, ultrafast video via rolling-shutter coding~\cite{antipa2019video}, and privacy-preserving imaging~\cite{satat2017object, henry2023privacy}---making them appealing for embedded vision under strict size and cost constraints~\cite{kim2024high,ge2024lpsnet, xiangjun2025reveal}.

Image reconstruction becomes particularly challenging when the optical system produces highly multiplexed measurements, often modeled as 2D or 3D convolutions. In such cases, extended PSFs distribute scene information broadly across the sensor, leading to loss of spatial detail and strong overlap between measurements, which makes inversion ill-posed. Similar challenges arise in a range of computational imaging settings, from conventional cameras under severe aberrations or motion blur to tasks such as imaging through scattering media~\cite{yoon2020deep}, non-line-of-sight imaging~\cite{faccio2020non}, coherent diffractive imaging~\cite{miao2015beyond}, and microscopy with engineered PSFs~\cite{pavani2009three}.

We begin with a baseline shift-invariant model, where the system response is identical across locations and the measurement is a 2D convolution between the scene and a static PSF:
\begin{equation}
y[i,j] \;=\; \sum_{a,b} h[a,b]\, x[i-a,j-b] \;+\; \eta[i,j],
\label{eq:si}
\end{equation}
where $x,y \in \mathbb{R}^{H \times W}$ denote the scene irradiance and the captured measurement, $h$ is a static PSF of the system, and $\eta$ models additive noise.

In practice, most imaging systems are not truly shift-invariant, even though they are often modeled as convolutions. Off-axis aberrations, depth-dependent
propagation, field-dependent magnification, vignetting or pupil clipping, and
sensor truncation all make the effective system response depend on spatial
location~\cite{booth2014adaptive,thiebaut2016spatially,antipa2017diffusercam}.
This is especially pronounced for phase or coded masks with high effective numerical
aperture: resolution improves on-axis, but aberration-induced shift variance grows with
field angle. As a result, the location-dependent PSF $h_{i,j}$ at pixel position $[i,j]$ widens, skews, or changes phase structure
across the field, necessitating a spatially varying model. A more general shift-variant
model accounts for this effect by allowing the PSF to vary with the output coordinates:
\begin{equation}
y[i,j] \;=\; \sum_{a,b} h_{i,j}[a,b]\, x[i-a,j-b] \;+\; \eta[i,j],
\label{eq:sv}
\end{equation}
The forward model is no longer a 2D convolution but a large, spatially varying operator, raising the computational and memory cost of precise inversion; sensor cropping and noise further increase ill-posedness and calibration sensitivity.

\subsection{Image Restoration}
Given such forward models, image recovery in lensless cameras is carried out through computational inversion, often posed as deconvolution, closely related to reconstruction in conventional cameras where strong degradation can arise from aberrations, motion, or turbulence. Classical estimators such as Wiener filtering~\cite{wiener1964extrapolation} and Richardson--Lucy iterations~\cite{richardson1972bayesian, lucy1974iterative} provide long-standing baselines and are attractive when the forward model is accurate, but their practical performance is often limited by noise amplification and PSF misspecification. To handle non-ideal forward models that include truncation/cropping and non-smooth priors (e.g., total variation, TV), iterative optimization frameworks such as ADMM~\cite{boyd2011distributed} are widely used to decouple data fidelity from regularization and enable tractable solvers~\cite{antipa2017diffusercam}.

In parallel, deep networks, including CNN-based restorers~\cite{ronneberger2015u,zhang2018image,bae2020lensless,chen2022simple} and ViT architectures~\cite{dosovitskiy2020image,pan2022image}, learn direct mappings from measurements to images and have become strong empirical baselines for large-scale restoration. Motivated by their complementary strengths, a growing line of work integrates explicit forward-model structure into learning-based pipelines. Examples include reconstruction via unrolled iterations with learned denoisers~\cite{monakhova2019learned, kingshott2022unrolled, poudel2024deeplir, bezzam2025towards}, measurement-consistency constraints for unsupervised training~\cite{ulyanov2018deep, wang2020phase, monakhova2021untrained}, or feed-forward hybrids that combine a physical inversion stage with learned refinement~\cite{khan2020flatnet, yanny2022deep}. Related approaches embed deconvolution within multi-scale feature hierarchies to improve fidelity and robustness~\cite{dong2021dwdn, li2023mwdns, bai2025lensnet}. 
The above methods differ primarily in how they use the measurements: many enforce physics through a fixed forward-model constraint or through a single inversion step, after which the measurement-domain information is no longer explicitly carried through the hierarchy. Since inversion difficulty depends strongly on the PSF support and the conditioning of the forward operator, we next formalize the forward model and summarize three fundamental failure modes that emerge as the PSF support grows and model mismatch increases. This motivates architectures that preserve and update measurement- and image-domain representations in tandem.

\section{Key Challenges}
\label{sec:challenges}
We consider a linear forward model $y_\gamma = H_\gamma x + \eta$, where $x \in \mathbb{R}^N$ is the scene, $y_\gamma \in \mathbb{R}^M$ the measurement, $H_\gamma \in \mathbb{R}^{M \times N}$ a forward operator parameterized by complexity $\gamma$, and $\eta$ sensor noise. As $\gamma$ grows, the PSF widens and mixes information across increasingly distant pixels, creating three failure modes for neural restoration that motivate the design of IFIN.

\subsubsection{Locality Mismatch Under Degradation.}
A wide-support $H_\gamma$ aggregates long-range contributions into each measurement, whereas CNNs and ViTs process $y_\gamma$ with finite receptive fields or windowed self-attention~\cite{luo2016erf, liu2021swin}. When the kernel support exceeds a layer's effective field, out-of-field contributions behave as structured noise, amplifying the difficulty posed by $H_\gamma$ through the mismatch between its large footprint and the network's locality.

\subsubsection{Loss from Model-Based Inversion.}
A second source of degradation arises from how the inverse problem is solved. When $H_\gamma$ is ill-conditioned, the consistent set $\mathcal{S}(y_\gamma)=\{\, x \in \mathbb{R}^N : H_\gamma x \approx y_\gamma \,\}$ is high-dimensional, so weakly constrained high-frequency directions are nearly unobservable and classical or learned inverses resolve this by selecting a single smooth estimate $\hat{x} = G(y_\gamma)$~\cite{bertero2021introduction,chen2020deep}. One-sided physics--NN pipelines that feed only $\hat{x}$ into a network~\cite{monakhova2019learned,khan2020flatnet,kingshott2022unrolled,yanny2022deep,poudel2024deeplir,bezzam2025towards} discard the measurement-domain residuals that show how $\hat{x}$ fails to explain $y_\gamma$, leaving fine detail to learned priors rather than data consistency.

\subsubsection{Representation Bottlenecks.}
Even with inversion embedded in dimension-reducing encoders $z_\gamma = f_\theta(y_\gamma)$~\cite{dong2021dwdn, li2023mwdns, bai2025lensnet}, non-invertibility gives $I(x; z_\gamma) \le I(x; y_\gamma)$~\cite{cover1999elements}, and as $\gamma$ grows the increasingly ill-conditioned, low-rank $H_\gamma$ pushes more fine-detail components below the encoder's effective thresholds. Inserting inversion alone is therefore insufficient: unless the architecture explicitly preserves and propagates measurement information across layers, it remains vulnerable to the combined effects of ill-conditioned low-rank forward models and representation bottlenecks.

\section{Method}
\label{sec:method}

\begin{figure}[t]
  \centering
  \includegraphics[width=\linewidth]{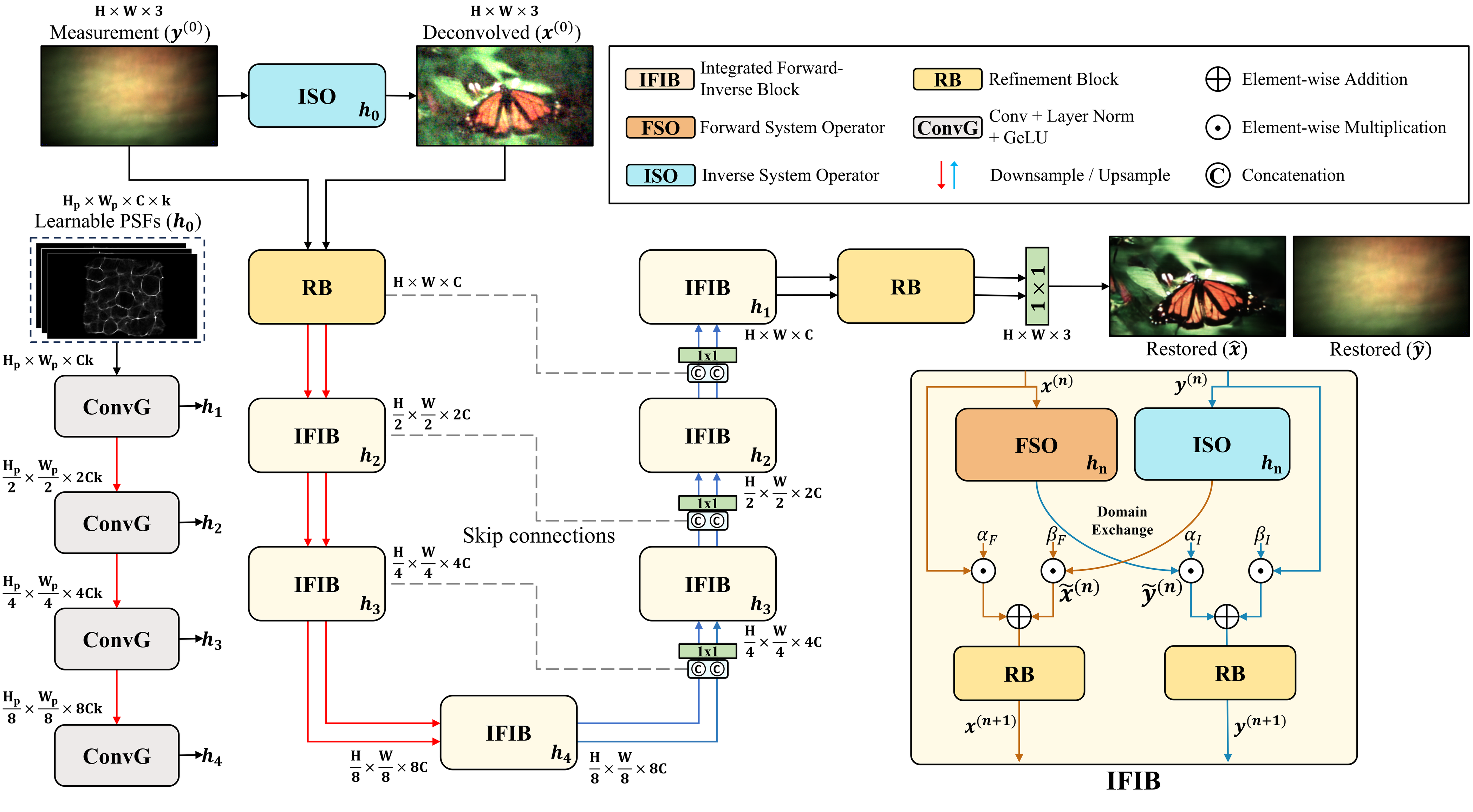}
  \caption{Overall architecture of IFIN. The network follows an encoder--decoder structure, where Integrated Forward--Inverse Blocks (IFIBs) are inserted at each scale to jointly apply the Forward System Operator (FSO) and Inverse System Operator (ISO). A shared learnable PSF field guides both operators, ensuring forward--inverse consistency across scales. Notation: $n$ indexes the per-scale PSF embedding $h_n$ (one per scale, shared across the hierarchy); $(n)$ denotes the stage-$n$ representations ($x^{(n)}, y^{(n)}$) updated across the hierarchy.}
  \label{fig:overview}
\end{figure}

The proposed IFIN adopts an encoder--decoder backbone equipped with
Integrated Forward--Inverse Blocks (IFIBs) at every scale
(\cref{fig:overview}). The encoder progressively downsamples the input measurement and coarse estimation to capture coarse-scale coupling and long-range interactions, while the decoder upsamples features back to the native resolution to recover fine details. At each resolution, an IFIB couples
a Forward System Operator (FSO), which maps image-domain features to the
measurement domain, with an Inverse System Operator (ISO), which restores
image-domain features from the measurement. Across scales, both operators condition on and jointly refine a learnable PSF field, enforcing forward--inverse consistency while propagating physically meaningful residual signals throughout the network.

At the input stage, a coarse estimation is obtained by applying
the ISO to the measurement. The pair \emph{(measurement, coarse estimation)}
is then propagated as two coupled streams through the encoder--decoder
hierarchy. Within each IFIB, the FSO and ISO exchange features
bidirectionally, jointly enforcing consistency across the measurement
and image domains.

\subsection{Learnable PSF Field}
IFIN incorporates a learnable PSF representation that provides explicit system awareness to both the FSO and the ISO. The PSF field is parameterized as $k{=}s^2$ kernels covering local regions of the image. When $s{=}1$, the PSF field reduces to a single global kernel. Kernels can be initialized from calibrated measurements, a single reference PSF, or random patterns, and are jointly optimized end-to-end within the network. In our experiments we use $k\in\{1,4,9,16\}$ for DiffuserCam and $k{=}9$ for WiderCam and MultiWienerNet (MWNet), with cropped PSF supports of $270\times270$, $135\times135$, and $320\times224$ for the three lensless benchmarks (and $101\times101$ for Gaussian deblurring).

A compact PSF encoder maps the field to multi-scale embeddings $\{h_n\}$, where each $h_n$ is shared between the encoder and decoder IFIBs at scale $n$ and injected into both the FSO and the ISO, maintaining physical consistency across the hierarchy. Because the same PSF field must explain both how features generate measurements (FSO) and how measurements invert to sharp images (ISO), the PSFs are constrained by complementary supervision in both domains, which improves identifiability, discourages degenerate kernels, and aids blind PSF estimation.

We normalize the PSF to obtain unit DC gain in both the FSO and the ISO. This stabilizes the physics operators and prevents scale drift. We further impose a weak non-negativity regularizer on the PSF by penalizing negative values during training, guiding the learned kernels toward physically plausible solutions.

\subsection{Integrated Forward--Inverse Block (IFIB)}
\begin{figure}[t]
  \centering
  \includegraphics[width=\linewidth]{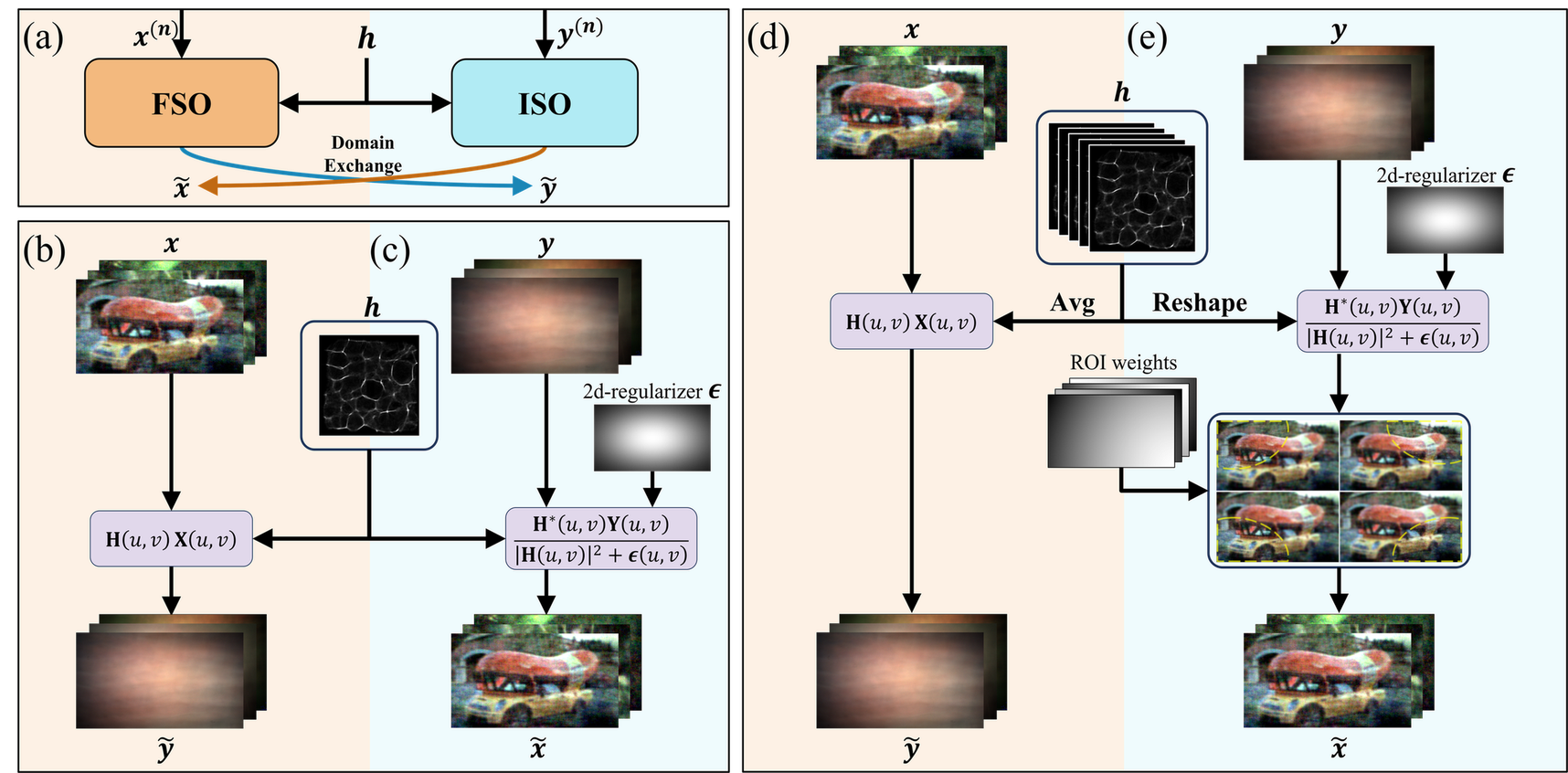}
    \caption{(a) Schematic of the forward--inverse pairing in the IFIB. (b,c) \textbf{Single-PSF setting:} FSO uses 2D convolution; ISO uses Wiener-like deconvolution. (d,e) \textbf{PSF-field setting:} FSO uses a single representative (averaged) PSF from the PSF field; ISO applies region-wise deconvolution blended by learnable region-of-interest (ROI) maps.}
  \label{fig:FSOISO}
\end{figure}

The IFIB is the fundamental unit of IFIN, coupling forward and inverse imaging at each scale through two parallel operators, FSO and ISO (\cref{fig:FSOISO}), both tied to the system's physics. They can be configured as shift-invariant when degradations are approximately uniform, or spatially varying for more complex degradations, letting IFIN balance efficiency and fidelity.

\subsubsection{Forward System Operator (FSO).}
The FSO simulates how the current image-domain estimate $x$ would be formed by
the imaging system, producing a forward projection in the measurement domain.
By default, we use 2D linear convolution with zero padding via a single
PSF $h$:
\begin{equation}
\tilde{y}[i,j] \;=\; (x * h)[i,j].
\label{eq:FSO-conv}
\end{equation}

The projection $\tilde{y}$ is a measurement-domain proxy induced by the current image representation. Inside each IFIB it augments the measurement-domain features to improve their consistency with the forward model and to better support the subsequent inverse update. For efficiency under large-footprint PSFs, we compute it in the frequency domain with zero padding and crop back to the native resolution.
When the PSF field provides multiple kernels, the FSO uses a single representative (averaged) kernel for this projection. Because its role is to supply a stable measurement-consistency cue rather than to synthesize a high-fidelity measurement, an averaged kernel reduces model mismatch and avoids the cost of region-wise forward projection. In contrast, the ISO carries out the locally sensitive deconvolution and therefore applies region-wise inversion with ROI blending (below) to capture field-dependent degradation.

\subsubsection{Inverse System Operator (ISO).}
The ISO restores a sharp estimate from the degraded measurement via Wiener-like
deconvolution with a learnable frequency-dependent regularizer.
For PSF $h$, letting $Y(u,v)=\mathcal{F}\!\{\,W\cdot P_{rp} y\,\}(u,v)$ and
$H(u,v)=\mathcal{F}\{h\}(u,v)$, where $\mathcal{F}\{\cdot\}$ denotes the Fourier transform,
$P_{rp}$ is replicate padding, and $W$ is a mild Gaussian window used to mitigate
wrap-around artifacts during deconvolution~\cite{khan2020flatnet}, we compute:
\begin{equation}
\widehat{X}(u,v) \;=\; \frac{H^*(u,v)}{|H(u,v)|^2 + \epsilon(u,v)} \, Y(u,v),
\qquad \epsilon(u,v) \ge 0,
\label{eq:ISO-si}
\end{equation}
and set $\hat{x}=\mathcal{F}^{-1}\{\widehat{X}\}$.
Here, $\epsilon(u,v)$ is a learnable 2D parameterization refined during training,
with non-negativity enforced by a ReLU. After the inverse Fourier transform, we crop the result back to the native resolution, matching the forward projection size.
Within each IFIB, this reconstructed estimate is forwarded as an augmentation
signal to the subsequent image-stream refinement.

When the PSF field provides multiple kernels, we can apply the same inverse \cref{eq:ISO-si} region-wise using PSFs $\{h_r\}_{r=1}^k$ and
regularizers $\{\epsilon_r\}_{r=1}^k$, and blend the reconstructions
with normalized ROI weights to capture local variability:
\begin{equation}
\hat{x}[i,j] \;=\; \sum_{r=1}^{k} w_r[i,j] \, \mathcal{F}^{-1}\!\{\widehat{X}_r\}[i,j],
\label{eq:ISO-sv-blend}
\end{equation}
where $\widehat{X}_r$ denotes the result of \cref{eq:ISO-si} computed with the region-wise pair $(h_r,\epsilon_r)$, and $\{w_r\}_{r=1}^k$ are $k$ learnable ROI maps. We initialize the ROI maps from
Gaussian kernels $\{g_r\}_{r=1}^{k}$ centered at $p_r$, where $k{=}s^2$ and $p_r$ are
the centers of an $s\times s$ grid partitioning the input measurement:
\begin{equation}
g_r[i,j] \;=\; \exp\!\Big(-\tfrac{\|(i,j)-p_r\|_2^2}{2\sigma_r^2}\Big),
\qquad
w_r[i,j] \;=\; \frac{g_r[i,j]}{\sum_{q=1}^{k} g_q[i,j]},
\end{equation}
\begin{equation}
\sum_{r=1}^{k} w_r[i,j] \;=\; 1 \ \ \forall (i,j),
\end{equation}
where $\sigma_r$ is the width of the $r$-th initialization Gaussian.

\subsubsection{Integrated Forward--Inverse.}
\label{sec:ifi}
The hallmark of the IFIB is the bidirectional exchange between FSO and ISO.
At each stage ${(n)}$, the FSO produces a measurement-domain projection
$\tilde{y}^{(n)}=\mathrm{FSO}(x^{(n)};h_n)$ from the current image-domain
representation, while the ISO produces an image-domain estimate (a feature map)
$\tilde{x}^{(n)}=\mathrm{ISO}(y^{(n)};h_n)$ from the current
measurement-domain representation. These cross-domain outputs are then used as
augmentation signals that are fused into the next updates of both streams:
\begin{equation}
y^{(n+1)}=\phi_{\theta}^y\!\left(\alpha_F^{(n)}\cdot y^{(n)}+\beta_F^{(n)}\cdot \tilde{y}^{(n)}\right),
\end{equation}
\begin{equation}
x^{(n+1)}=\phi_{\theta}^x\!\left(\alpha_I^{(n)}\cdot x^{(n)}+\beta_I^{(n)}\cdot \tilde{x}^{(n)}\right),
\end{equation}
where $\phi_{\theta}^{y}$ and $\phi_{\theta}^{x}$ each consist of a sequence of convolutional layers, and
$\alpha_F^{(n)},\beta_F^{(n)}$ and $\alpha_I^{(n)},\beta_I^{(n)}$ are learnable per-channel scalar gates for the measurement and image streams at scale $n$.
Thus $\tilde{y}^{(n)}$ updates the measurement representation before the next ISO step, while $\tilde{x}^{(n)}$ drives the image-domain update and conditions the next FSO projection, forming a forward--inverse coupling across the hierarchy. Layer configurations and comprehensive FSO/ISO ablations are in the supplementary material.

\section{Results}
We evaluate IFIN via end-to-end supervised training on paired scene--measurement data spanning real display--capture and synthetic settings. We report results on three lensless benchmarks---\textbf{DiffuserCam}~\cite{monakhova2019learned}, \textbf{WiderCam} (ours), and \textbf{MultiWienerNet (MWNet)}~\cite{yanny2022deep}---which together cover a widely used diffuser dataset, strong shift variance with a large field of view, and multi-scale PSF fields with real experimental validation. We compare against classical baselines~\cite{boyd2011distributed, wiener1964extrapolation}, data-driven models~\cite{chen2022simple, ronneberger2015u}, and representative hybrid approaches~\cite{monakhova2019learned, poudel2024deeplir, yanny2022deep, kingshott2022unrolled, li2023mwdns, bai2025lensnet, bezzam2025towards}. To probe the scope of IFIN beyond lensless imaging, we additionally include synthetic \textbf{Gaussian deblurring} and \textbf{diffractive reconstruction} benchmarks. All methods use a unified reconstruction objective with standard method-specific recipes; further dataset, implementation, and training details are in the supplementary material.

\subsubsection{DiffuserCam.}
DiffuserCam~\cite{monakhova2019learned} contains 25{,}000 paired display--capture measurements acquired with the DiffuserCam prototype~\cite{antipa2017diffusercam} (24{,}000 train / 1{,}000 test); we use the standard $480\times270$ measurements obtained by $4\times$ downsampling from $1920\times1080$ raw sensor frames.

\begin{figure}[t]
  \centering
  \includegraphics[width=\linewidth]{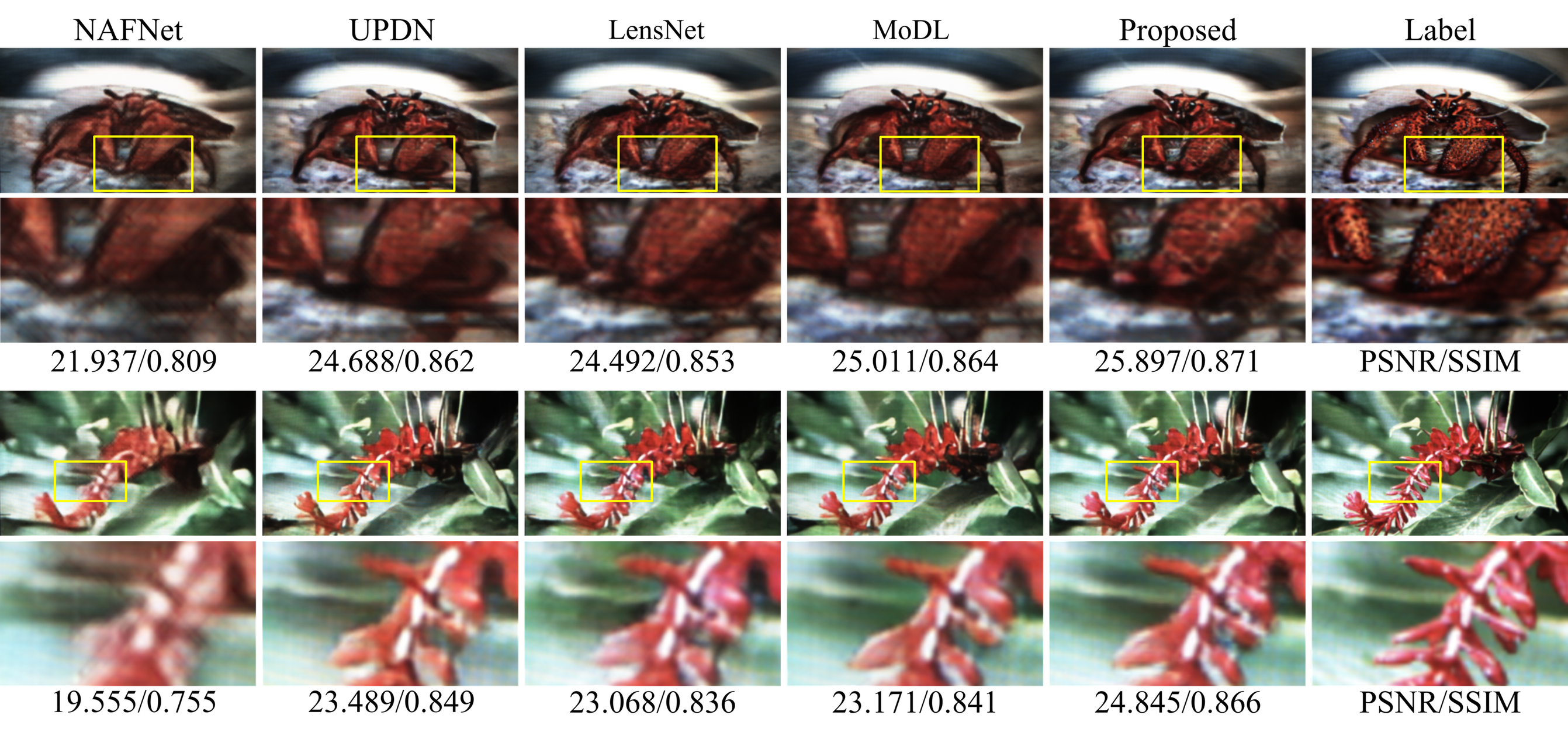}
  \caption{Visual comparison on DiffuserCam display--capture data. IFIN preserves color fidelity and high-frequency textures while suppressing artifacts. Insets mark zoomed regions and structures.}
  \label{fig:result_diffusercam}
\end{figure}

On this benchmark, IFIN improves PSNR, LPIPS, and SSIM over prior methods (\cref{tab:quantitative}) and yields clearer reconstructions with fewer ringing/haze artifacts while better preserving textures and color consistency (\cref{fig:result_diffusercam}).
These results are consistent with the design rationale of IFIN, which repeatedly reintroduces measurement-consistency cues during decoding via forward--inverse refinement.

\begin{table}[!t]
\centering
\caption{
Quantitative comparison on three benchmarks---\textbf{DiffuserCam}, \textbf{WiderCam}, and \textbf{MWNet dataset}. \textbf{ISO (Ours)} denotes the learned initial inverse operator used for the coarse estimation stage of IFIN.
We report PSNR~$\uparrow$, LPIPS~$\downarrow$~\cite{zhang2018unreasonable}, and SSIM~$\uparrow$ (arrows indicate the preferred direction).
Best in bold, second best underlined.
}
\label{tab:quantitative}
\resizebox{\linewidth}{!}{
\begin{tabular}{l ccc ccc ccc}
\toprule
Dataset & \multicolumn{3}{c}{DiffuserCam} & \multicolumn{3}{c}{WiderCam} & \multicolumn{3}{c}{MultiWienerNet} \\
\cmidrule(lr){1-1} \cmidrule(lr){2-4} \cmidrule(lr){5-7} \cmidrule(lr){8-10}
Metrics & PSNR $\uparrow$ & LPIPS $\downarrow$ & SSIM $\uparrow$
 & PSNR $\uparrow$ & LPIPS $\downarrow$ & SSIM $\uparrow$
 & PSNR $\uparrow$ & LPIPS $\downarrow$ & SSIM $\uparrow$ \\
\midrule
ADMM~\cite{boyd2011distributed}              & 12.252 & 0.607 & 0.346 & 11.843 & 0.643 & 0.323 & 19.189 & 0.557 & 0.420 \\
Wiener Deconv.~\cite{wiener1964extrapolation}    & 12.552 & 0.591 & 0.384 & 12.405 & 0.607 & 0.369 & 18.658 & 0.640 & 0.302 \\
\textbf{ISO (Ours)}      & 16.528 & 0.544 & 0.404 & 17.240 & 0.462 & 0.444 & 20.202 & 0.623 & 0.380 \\
U-Net~\cite{ronneberger2015u}              & 21.230 & 0.394 & 0.656 & 21.890 & 0.474 & 0.646 & 23.859 & 0.389 & 0.589 \\
NAFNet~\cite{chen2022simple}            & 24.830 & 0.239 & 0.810 & 23.857 & 0.245 & 0.769 & 24.657 & 0.282 & 0.712 \\
Le-ADMM-U~\cite{monakhova2019learned}         & 23.261 & 0.312 & 0.765 & 21.956 & 0.278 & 0.748 & 23.732 & 0.335 & 0.702 \\
DeepLIR~\cite{poudel2024deeplir}           & 25.958 & 0.260 & 0.829 & 20.523 & 0.339 & 0.642 & 22.556 & 0.379 & 0.642 \\
MWNet~\cite{yanny2022deep}             & 24.832 & 0.247 & 0.810 & 23.001 & 0.255 & 0.766 & 25.660 & 0.260 & 0.728 \\
UPDN~\cite{kingshott2022unrolled}              & \underline{28.228} & 0.194 & 0.877 & 23.920 & 0.229 & 0.801 & 24.364 & 0.287 & 0.707 \\
MWDNs~\cite{li2023mwdns}              & 27.298 & 0.217 & 0.845 & 24.525 & 0.224 & 0.801 & 27.436 & 0.236 & 0.780 \\
LensNet~\cite{bai2025lensnet}           & 27.650 & 0.201 & 0.868 & 24.615 & 0.219 & 0.806 & 27.546 & 0.221 & 0.809 \\
MoDL~\cite{bezzam2025towards}              & 27.958 & \underline{0.183} & \underline{0.878} & \underline{24.791} & \underline{0.202} & \underline{0.810} & \underline{28.504} & \underline{0.202} & \underline{0.831} \\
\textbf{IFIN (Ours)} & \textbf{29.862} & \textbf{0.174} & \textbf{0.893}
                  & \textbf{25.444} & \textbf{0.201} & \textbf{0.824}
                  & \textbf{31.083} & \textbf{0.175} & \textbf{0.866} \\
\bottomrule
\end{tabular}}
\end{table}

\begin{figure}[t]
  \centering
  \includegraphics[width=\linewidth]{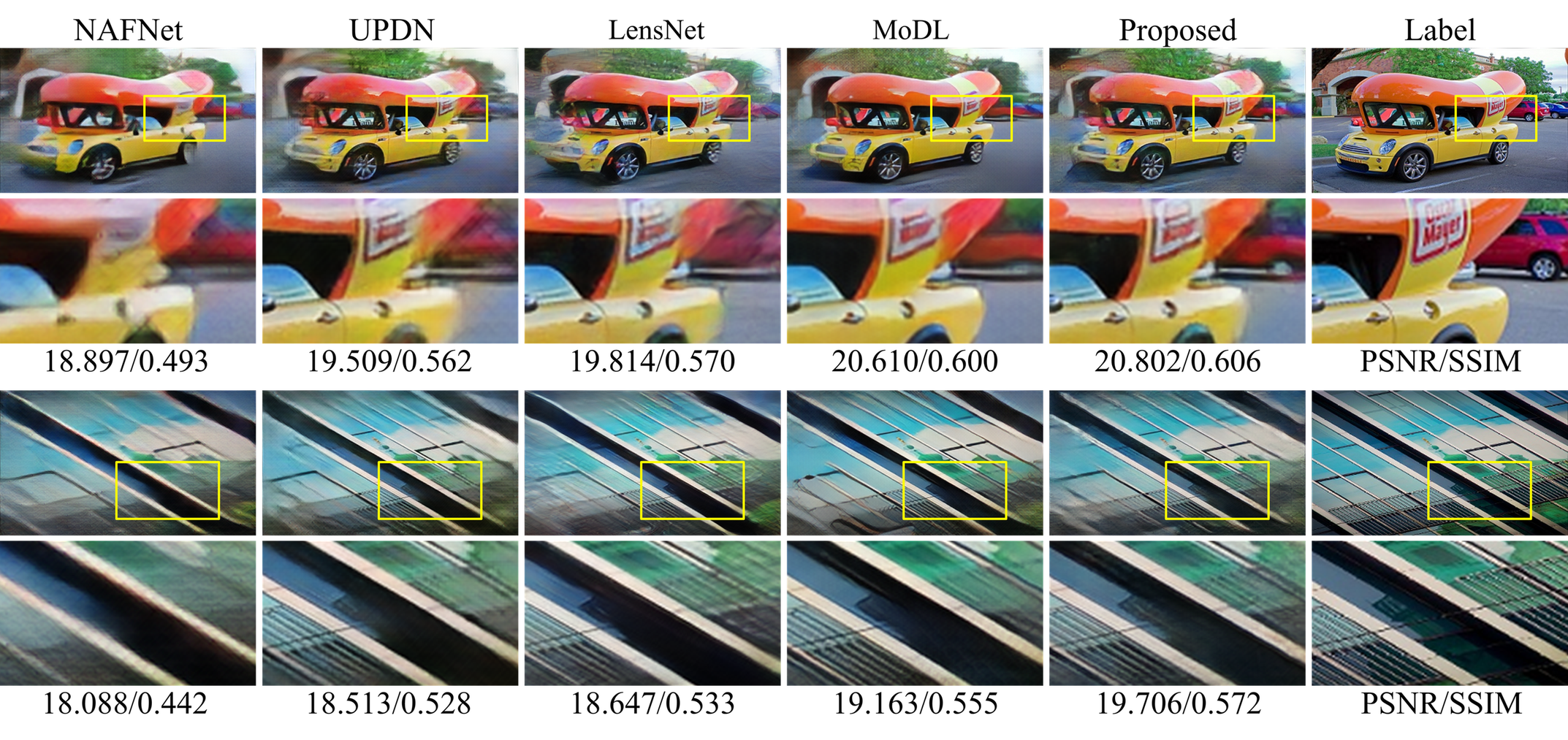}
  \caption{Comparison on WiderCam dataset. Compared to prior methods, IFIN mitigates field-dependent peripheral blur and geometric distortion, while preserving fine textures and edges across the entire image.}
  \label{fig:result_SV}
\end{figure}

\begin{figure}[!t]
  \centering
  \includegraphics[width=\linewidth]{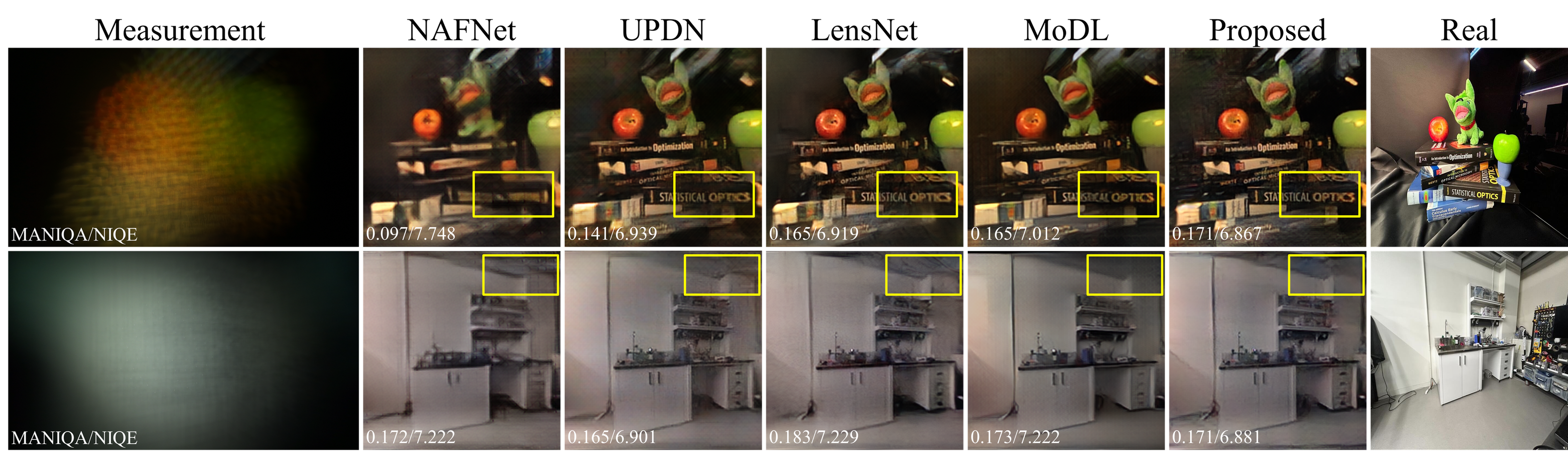}
  \caption{In-the-wild WiderCam measurements. IFIN generalizes to diverse scenes and lighting, reducing ringing and color shifts while preserving edges and textures. Without ground truth, we additionally report no-reference image quality metrics: MANIQA$\uparrow$~\cite{yang2022maniqa} and NIQE$\downarrow$~\cite{mittal2012making}.}
  \label{fig:result_inthewild}
\end{figure}

\subsubsection{WiderCam.}
WiderCam (ours) consists of 25{,}000 wide field-of-view (FoV, $>100^\circ$) lensless measurements (24{,}000 train / 1{,}000 test), resized to $480\times270$ from $4608\times2592$ sensor frames, with affine-aligned supervision.
The dataset exhibits strong field dependence: regions near the optical center are typically easier to reconstruct, while peripheral regions are substantially more challenging due to shift variance and non-uniform degradation.

IFIN shows its advantage most clearly in the outer field, recovering sparse peripheral structures more reliably while maintaining sharp central content (\cref{fig:result_SV}), consistent with the quantitative gains in \cref{tab:quantitative}.
We further test generalization on in-the-wild captures without ground truth (\cref{fig:result_inthewild}); IFIN remains stable across diverse scenes and illumination, reducing ringing and color shifts while preserving edge fidelity.

\subsubsection{MWNet Dataset.}
\begin{figure}[t]
  \centering
  \includegraphics[width=\linewidth]{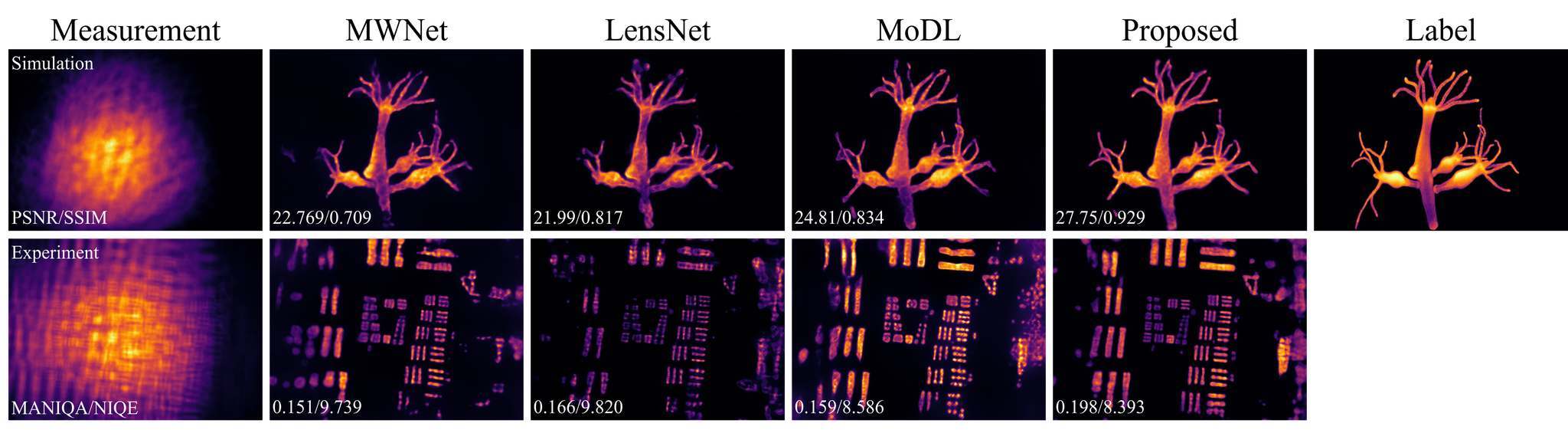}
  \caption{Comparison on MWNet dataset. The first row shows simulated spatially varying measurements and reconstructions; the second row shows experimental miniscope captures of a USAF target.}
  \label{fig:result_MW}
\end{figure}

MultiWienerNet (MWNet)~\cite{yanny2022deep} provides 22{,}125 paired 2D samples (17{,}700 train / 4{,}425 test) generated using a measured spatially varying PSF field from a mask-based miniscope~\cite{yanny2020miniscope3d}; in our experiments, we resize images to $320\times224$ for efficiency.
On this benchmark, IFIN improves quantitative metrics (\cref{tab:quantitative}) and produces visibly sharper reconstructions from both synthetic measurements and real USAF captures (\cref{fig:result_MW}).
IFIN does not require explicit region-wise PSF initialization: starting from limited calibration (e.g., a single on-axis PSF), it can learn effective position-dependent behavior during reconstruction.
Compared to MWNet, which relies on multiple calibrated PSFs (9 in the original setting) to recover off-axis information, IFIN surpasses it by a clear margin while requiring far less calibration. It exceeds MWNet without the learnable PSF field and even without calibrated PSF information, while learning the PSF field end-to-end improves performance further.

\subsubsection{Gaussian Deblurring Simulation.}
In addition to lensless experiments, we evaluate IFIN on synthetic Gaussian deblurring (\cref{tab:deblur}).
We blur $384\times384$ images with Gaussian kernels of increasing standard deviation $\sigma$ and compare against strong purely data-driven restorers~\cite{chen2022simple, zhang2018image}.

For mild blur, NAFNet/RCAN remain highly competitive, yet IFIN achieves comparable performance, indicating that the forward--inverse interleaving prior is not limited to large-footprint degradations.
As the blur footprint increases, IFIN becomes increasingly advantageous and yields the best overall performance at $\sigma=15$ (PSNR/SSIM$\uparrow$, LPIPS$\downarrow$).
Overall, these results suggest that incorporating explicit forward and inverse operators helps IFIN degrade more gracefully as the inverse problem becomes more ill-conditioned, while remaining effective in the small-footprint regime.

\subsubsection{Diffractive Imaging Reconstruction.}
We evaluate whether IFIN transfers to an inverse problem with a fundamentally different forward operator by testing simulated inline holography, a diffractive imaging modality governed by wave propagation.
Using the Dogs vs.\ Cats dataset, we generate holographic measurements from $256\times256$ images via the angular spectrum method at $z=30$\,mm and $\lambda=532$\,nm, add a mixture of Gaussian and Poisson noise, and reconstruct the amplitude-only transmission object.

To adapt IFIN to inline holography, we keep the interleaving backbone and replace the system operators with modality-matched ones:
the FSO is instantiated with forward propagation using the angular spectrum method, and the ISO uses the corresponding back-propagation operator as an inversion module.
Since this setting is governed by wave propagation rather than a spatial PSF field, we omit the PSF encoder and do not perform PSF-field conditioning in this experiment. A detailed description of the dataset and the corresponding adaptations is provided in the supplementary material, with additional qualitative results.

Despite these operator-level changes, IFIN outperforms purely data-driven baselines~\cite{chen2022simple, zhang2018image} and a physics-initialized baseline that applies a single back-propagation step followed by NAFNet refinement (NAFNet+) (\cref{tab:holo}).
This result indicates that the interleaving mechanism is transferable: when the forward and inverse operators are swapped to match a different imaging physics, the same architectural principle continues to provide effective guidance beyond deconvolution.

\begin{table*}[t]
\centering
\begin{minipage}[b]{0.665\textwidth}
\centering
\caption{Gaussian deblurring with standard deviations $\sigma \in \{5,10,15\}$. We report PSNR~$\uparrow$/LPIPS~$\downarrow$/SSIM~$\uparrow$.}
\label{tab:deblur}
\begin{adjustbox}{max width=\linewidth}
\begin{tabular}{lccccccccc}
\toprule
\multirow{2}{*}{\textbf{Method}} 
& \multicolumn{3}{c}{$\sigma=5$} 
& \multicolumn{3}{c}{$\sigma=10$} 
& \multicolumn{3}{c}{$\sigma=15$} \\
\cmidrule(lr){2-4}\cmidrule(lr){5-7}\cmidrule(lr){8-10}
& PSNR $\uparrow$ & LPIPS $\downarrow$ & SSIM $\uparrow$ 
& PSNR $\uparrow$ & LPIPS $\downarrow$ & SSIM $\uparrow$ 
& PSNR $\uparrow$ & LPIPS $\downarrow$ & SSIM $\uparrow$ \\
\midrule
RCAN & \underline{25.303} & \underline{0.220} & \underline{0.810}
     & 22.623 & \underline{0.330} & \underline{0.750}
     & 21.280 & 0.355 & 0.711 \\
NAFNet & \textbf{25.625} & \textbf{0.205} & \textbf{0.825}
       & \textbf{22.810} & \textbf{0.315} & \textbf{0.766}
       & \underline{21.535} & \underline{0.345} & \underline{0.715} \\
\textbf{IFIN (Ours)} & 25.100 & 0.230 & 0.800
                  & \underline{22.732} & 0.337 & 0.741
                  & \textbf{21.654} & \textbf{0.340} & \textbf{0.721} \\
\bottomrule
\end{tabular}
\end{adjustbox}
\end{minipage}
\hfill
\begin{minipage}[b]{0.295\textwidth}
\centering
\caption{Inline holography ($z=30$\,mm; $\lambda=532$\,nm).}
\label{tab:holo}
\renewcommand{\arraystretch}{1.08}
\begin{adjustbox}{max width=\linewidth}
\begin{tabular}{lccc}
\toprule
\textbf{Method} & PSNR $\uparrow$ & LPIPS $\downarrow$ & SSIM $\uparrow$ \\
\midrule
RCAN & 23.764 & 0.354 & 0.702 \\
NAFNet & 23.224 & 0.389 & 0.634 \\
NAFNet+ & \underline{25.751} & \underline{0.242} & \underline{0.817} \\
\textbf{IFIN (Ours)} & \textbf{28.302} & \textbf{0.166} & \textbf{0.890} \\
\bottomrule
\end{tabular}
\end{adjustbox}
\end{minipage}

\end{table*}
\subsubsection{Ablation Study and Computational Cost.}
We isolate each component and quantify efficiency on DiffuserCam (\cref{tab:main_ablation} and \cref{tab:main_cost}; full ablations in the supplementary material). Replacing both operators with identities keeps the dual-stream paths but removes the physics, and the large drop shows that the gain originates in the forward--inverse coupling, not the extra pathways or parameters. Retaining a single direction (\emph{FSO-only} or \emph{ISO-only}) recovers only part of the gap, so the operators supply \emph{complementary} cues: measurement-domain consistency from one and image-domain recoverability from the other.

The other controls locate where this benefit arises. Removing the learnable 2D regularizer or replacing the refinement block (RB) with a plain convolutional gate (ConvG) lowers quality, so the physics operators pay off only when paired with a learnable stabilizer and enough capacity; dropping the initial ISO barely changes the result; the initial ISO mainly helps stabilize training. Scaling the PSF field ($k{=}1{\to}16$) yields consistent but saturating gains, since a few well-placed kernels already capture most of the shift variance, and a learned field surpasses frozen calibrated PSFs, indicating that end-to-end refinement absorbs residual calibration mismatch.

These gains do not stem from scale alone: IFIN ($k{=}1$) already surpasses every baseline at comparable parameters, FLOPs, and memory, so the improvement is architectural. Its overhead is the latency of the padded FFT-based FSO/ISO invoked at every block, making $k{=}1$ a competitive low-cost operating point and larger $k$ a deliberate accuracy-for-compute trade-off under strong shift variance.

\begin{table}[t]
\centering
\begin{minipage}[b]{0.41\linewidth}
\centering
\caption{Component and PSF-field ablation on DiffuserCam, plus learned-vs.-calibrated PSFs on MWNet ($k{=}9$).}
\label{tab:main_ablation}
\resizebox{\linewidth}{!}{
\begin{tabular}{lccc}
\toprule
\textbf{Setting} & PSNR\,$\uparrow$ & LPIPS\,$\downarrow$ & SSIM\,$\uparrow$ \\
\midrule
\multicolumn{4}{l}{\textit{Component controls --- DiffuserCam ($k{=}1$)}}\\
Identity FSO/ISO      & 24.674 & 0.255 & 0.800 \\
FSO-only (w/o ISO)    & 27.123 & 0.223 & 0.833 \\
ISO-only (w/o FSO)    & 28.711 & 0.185 & 0.882 \\
ConvG instead of RB   & 29.014 & 0.184 & 0.882 \\
w/o 2D regularizer    & 28.880 & 0.186 & 0.881 \\
w/o initial ISO       & 29.443 & \textbf{0.176} & 0.891 \\
\textbf{IFIN ($k{=}1$)}        & \textbf{29.535} & 0.179 & \textbf{0.892} \\
\addlinespace
\multicolumn{4}{l}{\textit{PSF-field scaling --- DiffuserCam}}\\
IFIN ($k{=}4$)        & 29.612 & 0.177 & 0.893 \\
IFIN ($k{=}9$)        & 29.751 & 0.176 & 0.893 \\
\textbf{IFIN ($k{=}16$)}       & \textbf{29.862} & \textbf{0.174} & \textbf{0.893} \\
\addlinespace
\multicolumn{4}{l}{\textit{Learned vs.\ calibrated PSF --- MWNet ($k{=}9$)}}\\
Frozen calibrated PSFs & 30.350 & 0.183 & 0.857 \\
\textbf{Learned PSF field}     & \textbf{31.083} & \textbf{0.175} & \textbf{0.866} \\
\bottomrule
\end{tabular}}
\end{minipage}\hfill
\begin{minipage}[b]{0.55\linewidth}
\centering
\caption{Computational cost on DiffuserCam (batch size 1, RTX A6000): parameters, FLOPs, peak VRAM, and inference time.}
\label{tab:main_cost}
\renewcommand{\arraystretch}{1.64}
\resizebox{\linewidth}{!}{
\begin{tabular}{lcccc}
\toprule
\textbf{Method} & Params (M) & FLOPs (G) & VRAM (GB) & Time (ms) \\
\midrule
NAFNet          & 29.2  & 32.4  & 0.257 & 24.65 \\
MWDNs            & 22.1  & 106.4 & 0.422 & 12.81 \\
UPDN            & 3.7   & 24.5  & 0.326 & 60.66 \\
LensNet         & 31.6  & 147.2 & 0.529 & 43.00 \\
MoDL            & 8.5   & 119.1 & 0.202 & 36.60 \\
\midrule
IFIN ($k{=}1$)  & 18.2  & 55.0  & 0.464 & 142.35 \\
IFIN ($k{=}4$)  & 66.2  & 62.4  & 1.257 & 156.33 \\
IFIN ($k{=}9$)  & 155.9 & 82.4  & 2.614 & 180.37 \\
IFIN ($k{=}16$) & 301.5 & 126.9 & 4.590 & 215.97 \\
\bottomrule
\end{tabular}}
\end{minipage}
\end{table}

\section{Conclusion}
We present \textbf{IFIN}, an integrated forward--inverse reconstruction architecture for lensless imaging that interleaves differentiable forward projections with learnable inverse updates across an encoder--decoder hierarchy. IFIN maintains coupled measurement- and image-domain streams and reintroduces measurement-consistency cues throughout decoding, letting the network refine image features while repeatedly checking their consistency under the imaging model. Furthermore, IFIN handles spatially varying systems and imperfect calibration using learnable PSF fields, allowing a single backbone to adapt to off-axis degradation that no stationary kernel can capture.

While IFIN achieves new state-of-the-art performance across diverse benchmarks (including our newly proposed WiderCam dataset) and shows promise in broader inverse problems, challenges remain. On the practical side, shift-variant systems with extended PSFs incur high per-block FFT overhead. The single-kernel ($k{=}1$) setting already offers a strong low-cost operating point, while denser PSF fields are reserved for stronger shift variance. IFIN also relies on a sufficiently stable forward model, as severe saturation or large geometry changes can erase information that no reconstructor can recover. We provide a comprehensive discussion of the generality, potential extensions, and limitations of our method in the supplementary material. Taken together, our results suggest that forward--inverse integration offers a practical template for designing reconstruction architectures in computational imaging systems.

\section*{Code and Data Availability}
Code, supplementary material, and the WiderCam dataset are available at
\url{https://iilab.io/IFIN/}.

\section*{Acknowledgements}
This work was supported in part by the National Research Foundation of Korea (NRF) grant funded by the Korean government (MSIT) (RS-2026-25495603); by the Technology Innovation Program (IRIS number: RS-2024-00419426, Development of light-electron beam based measurement and analysis instrument technologies for advanced packaging) funded by the Ministry of Trade, Industry \& Energy (MOTIE, Korea); and by the Creative-Pioneering Researchers Program through Seoul National University.

\bibliographystyle{splncs04}
\bibliography{main}

@String(ICCV  = {Int. Conf. Comput. Vis.})

@String(ECCV  = {Eur. Conf. Comput. Vis.})

@String(NeurIPS = {Adv. Neural Inform. Process. Syst.})

@String(ICASSP=	{ICASSP})

@String(ICCV  = {ICCV})

@String(ECCV  = {ECCV})

@String(NeurIPS = {NeurIPS})

@article{boominathan2020phlatcam,
  title={Phlatcam: Designed phase-mask based thin lensless camera},
  author={Boominathan, Vivek and Adams, Jesse K and Robinson, Jacob T and Veeraraghavan, Ashok},
  journal={IEEE transactions on pattern analysis and machine intelligence},
  volume={42},
  number={7},
  pages={1618--1629},
  year={2020},
  publisher={IEEE}
}

@article{mittal2012making,
  title={Making a “completely blind” image quality analyzer},
  author={Mittal, Anish and Soundararajan, Rajiv and Bovik, Alan C},
  journal={IEEE Signal processing letters},
  volume={20},
  number={3},
  pages={209--212},
  year={2012},
  publisher={IEEE}
}

@inproceedings{yang2022maniqa,
  title={Maniqa: Multi-dimension attention network for no-reference image quality assessment},
  author={Yang, Sidi and Wu, Tianhe and Shi, Shuwei and Lao, Shanshan and Gong, Yuan and Cao, Mingdeng and Wang, Jiahao and Yang, Yujiu},
  booktitle={Proceedings of the IEEE/CVF conference on computer vision and pattern recognition},
  pages={1191--1200},
  year={2022}
}

@inproceedings{chen2020deep,
  title={Deep decomposition learning for inverse imaging problems},
  author={Chen, Dongdong and Davies, Mike E},
  booktitle={European Conference on Computer Vision},
  pages={510--526},
  year={2020},
  organization={Springer}
}

@book{bertero2021introduction,
  title={Introduction to inverse problems in imaging},
  author={Bertero, Mario and Boccacci, Patrizia and De Mol, Christine},
  year={2021},
  publisher={CRC press}
}

@book{cover1999elements,
  title={Elements of information theory},
  author={Cover, Thomas M},
  year={1999},
  publisher={John Wiley \& Sons}
}

@article{bezzam2025towards,
  title={Towards Robust and Generalizable Lensless Imaging With Modular Learned Reconstruction},
  author={Bezzam, Eric and Perron, Yohann and Vetterli, Martin},
  journal={IEEE Transactions on Computational Imaging},
  volume={11},
  pages={213--227},
  year={2025},
  publisher={IEEE}
}

@article{bai2025lensnet,
  title={LensNet: An End-to-End Learning Framework for Empirical Point Spread Function Modeling and Lensless Imaging Reconstruction},
  author={Bai, Jiesong and Yin, Yuhao and Dong, Yihang and Zhang, Xiaofeng and Pun, Chi-Man and Chen, Xuhang},
  journal={arXiv preprint arXiv:2505.01755},
  year={2025}
}

@inproceedings{zhang2018image,
  title={Image super-resolution using very deep residual channel attention networks},
  author={Zhang, Yulun and Li, Kunpeng and Li, Kai and Wang, Lichen and Zhong, Bineng and Fu, Yun},
  booktitle={Proceedings of the European conference on computer vision (ECCV)},
  pages={286--301},
  year={2018}
}

@article{antipa2017diffusercam,
  title={DiffuserCam: lensless single-exposure 3D imaging},
  author={Antipa, Nick and Kuo, Grace and Heckel, Reinhard and Mildenhall, Ben and Bostan, Emrah and Ng, Ren and Waller, Laura},
  journal={Optica},
  volume={5},
  number={1},
  pages={1--9},
  year={2017},
  publisher={Optical Society of America}
}

@inproceedings{huiskes2008mir,
  title={The mir flickr retrieval evaluation},
  author={Huiskes, Mark J and Lew, Michael S},
  booktitle={Proceedings of the 1st ACM international conference on Multimedia information retrieval},
  pages={39--43},
  year={2008}
}

@article{faccio2020non,
  title={Non-line-of-sight imaging},
  author={Faccio, Daniele and Velten, Andreas and Wetzstein, Gordon},
  journal={Nature Reviews Physics},
  volume={2},
  number={6},
  pages={318--327},
  year={2020},
  publisher={Nature Publishing Group UK London}
}

@inproceedings{zhang2018unreasonable,
  title={The unreasonable effectiveness of deep features as a perceptual metric},
  author={Zhang, Richard and Isola, Phillip and Efros, Alexei A and Shechtman, Eli and Wang, Oliver},
  booktitle={Proceedings of the IEEE conference on computer vision and pattern recognition},
  pages={586--595},
  year={2018}
}

@article{yoon2020deep,
  title={Deep optical imaging within complex scattering media},
  author={Yoon, Seokchan and Kim, Moonseok and Jang, Mooseok and Choi, Youngwoon and Choi, Wonjun and Kang, Sungsam and Choi, Wonshik},
  journal={Nature Reviews Physics},
  volume={2},
  number={3},
  pages={141--158},
  year={2020},
  publisher={Nature Publishing Group UK London}
}

@inproceedings{sun2021loftr,
  title={LoFTR: Detector-free local feature matching with transformers},
  author={Sun, Jiaming and Shen, Zehong and Wang, Yuang and Bao, Hujun and Zhou, Xiaowei},
  booktitle={Proceedings of the IEEE/CVF conference on computer vision and pattern recognition},
  pages={8922--8931},
  year={2021}
}

@article{pavani2009three,
  title={Three-dimensional, single-molecule fluorescence imaging beyond the diffraction limit by using a double-helix point spread function},
  author={Pavani, Sri Rama Prasanna and Thompson, Michael A and Biteen, Julie S and Lord, Samuel J and Liu, Na and Twieg, Robert J and Piestun, Rafael and Moerner, William E},
  journal={Proceedings of the National Academy of Sciences},
  volume={106},
  number={9},
  pages={2995--2999},
  year={2009},
  publisher={National Academy of Sciences}
}

@article{miao2015beyond,
  title={Beyond crystallography: Diffractive imaging using coherent x-ray light sources},
  author={Miao, Jianwei and Ishikawa, Tetsuya and Robinson, Ian K and Murnane, Margaret M},
  journal={Science},
  volume={348},
  number={6234},
  pages={530--535},
  year={2015},
  publisher={American Association for the Advancement of Science}
}

@article{yanny2020miniscope3d,
  title={Miniscope3D: optimized single-shot miniature 3D fluorescence microscopy},
  author={Yanny, Kyrollos and Antipa, Nick and Liberti, William and Dehaeck, Sam and Monakhova, Kristina and Liu, Fanglin Linda and Shen, Konlin and Ng, Ren and Waller, Laura},
  journal={Light: Science \& Applications},
  volume={9},
  number={1},
  pages={171},
  year={2020},
  publisher={Nature Publishing Group UK London}
}

@inproceedings{chen2022simple,
  title={Simple baselines for image restoration},
  author={Chen, Liangyu and Chu, Xiaojie and Zhang, Xiangyu and Sun, Jian},
  booktitle={European conference on computer vision},
  pages={17--33},
  year={2022},
  organization={Springer}
}

@inproceedings{xiangjun2025reveal,
  title={Reveal Object in Lensless Photography via Region Gaze and Amplification},
  author={Xiangjun, Yin and Yue, Huihui},
  booktitle={The Thirteenth International Conference on Learning Representations},
  year={2025}
}

@article{booth2014adaptive,
  title={Adaptive optical microscopy: the ongoing quest for a perfect image},
  author={Booth, Martin J},
  journal={Light: Science \& Applications},
  volume={3},
  number={4},
  pages={e165--e165},
  year={2014},
  publisher={Nature Publishing Group}
}

@inproceedings{luo2016erf,
  title     = {Understanding the Effective Receptive Field in Deep Convolutional Neural Networks},
  author    = {Luo, Wenjie and Li, Yujia and Urtasun, Raquel and Zemel, Richard},
  booktitle = {NeurIPS},
  year      = {2016}
}

@inproceedings{liu2021swin,
  title     = {Swin Transformer: Hierarchical Vision Transformer Using Shifted Windows},
  author    = {Liu, Ze and Lin, Yutong and Cao, Yue and Hu, Han and Wei, Yixuan and Zhang, Zheng and Lin, Stephen and Guo, Baining},
  booktitle = {ICCV},
  year      = {2021}
}

@article{wang2020phase,
  title={Phase imaging with an untrained neural network},
  author={Wang, Fei and Bian, Yaoming and Wang, Haichao and Lyu, Meng and Pedrini, Giancarlo and Osten, Wolfgang and Barbastathis, George and Situ, Guohai},
  journal={Light: Science \& Applications},
  volume={9},
  number={1},
  pages={77},
  year={2020},
  publisher={Nature Publishing Group UK London}
}

@inproceedings{ronneberger2015u,
  title={U-net: Convolutional networks for biomedical image segmentation},
  author={Ronneberger, Olaf and Fischer, Philipp and Brox, Thomas},
  booktitle={International Conference on Medical image computing and computer-assisted intervention},
  pages={234--241},
  year={2015},
  organization={Springer}
}

@article{dosovitskiy2020image,
  title={An image is worth 16x16 words: Transformers for image recognition at scale},
  author={Dosovitskiy, Alexey and Beyer, Lucas and Kolesnikov, Alexander and Weissenborn, Dirk and Zhai, Xiaohua and Unterthiner, Thomas and Dehghani, Mostafa and Minderer, Matthias and Heigold, Georg and Gelly, Sylvain and others},
  journal={arXiv preprint arXiv:2010.11929},
  year={2020}
}

@article{asif2016flatcam,
  title={Flatcam: Thin, lensless cameras using coded aperture and computation},
  author={Asif, M Salman and Ayremlou, Ali and Sankaranarayanan, Aswin and Veeraraghavan, Ashok and Baraniuk, Richard G},
  journal={IEEE Transactions on Computational Imaging},
  volume={3},
  number={3},
  pages={384--397},
  year={2016},
  publisher={IEEE}
}

@inproceedings{thiebaut2016spatially,
  title={Spatially variant PSF modeling and image deblurring},
  author={Thi{\'e}baut, {\'E}ric and D{\'e}nis, Lo{\"\i}c and Soulez, Ferr{\'e}ol and Mourya, Rahul},
  booktitle={Adaptive Optics Systems V},
  volume={9909},
  pages={2211--2220},
  year={2016},
  organization={SPIE}
}

@article{sahoo2017single,
  title={Single-shot multispectral imaging with a monochromatic camera},
  author={Sahoo, Sujit Kumar and Tang, Dongliang and Dang, Cuong},
  journal={Optica},
  volume={4},
  number={10},
  pages={1209--1213},
  year={2017},
  publisher={Optical Society of America}
}

@article{satat2017object,
  title={Object classification through scattering media with deep learning on time resolved measurement},
  author={Satat, Guy and Tancik, Matthew and Gupta, Otkrist and Heshmat, Barmak and Raskar, Ramesh},
  journal={Optics express},
  volume={25},
  number={15},
  pages={17466--17479},
  year={2017},
  publisher={Optical Society of America}
}

@inproceedings{ge2024lpsnet,
  title={Lpsnet: End-to-end human pose and shape estimation with lensless imaging},
  author={Ge, Haoyang and Feng, Qiao and Jia, Hailong and Li, Xiongzheng and Yin, Xiangjun and Zhou, You and Yang, Jingyu and Li, Kun},
  booktitle={Proceedings of the IEEE/CVF Conference on Computer Vision and Pattern Recognition},
  pages={1471--1480},
  year={2024}
}

@inproceedings{kim2024high,
  title={High-speed lensless eye tracker for microsaccade measurement},
  author={Kim, Taeyoung and Lee, Kyung Chul and Lee, Kyungwon and Baek, Nakkyu and Jung, Jaewoo and Kim, Eosu and Park, Bobae and Ha, Junghee and Kim, Keun You and Seo, Young-Seok and others},
  booktitle={SPIE Advanced Biophotonics Conference (SPIE ABC 2023)},
  volume={13076},
  pages={38--44},
  year={2024},
  organization={SPIE}
}

@inproceedings{henry2023privacy,
  title={Privacy preserving face recognition with lensless camera},
  author={Henry, Chris and Asif, M Salman and Li, Zhu},
  booktitle={ICASSP 2023-2023 IEEE International Conference on Acoustics, Speech and Signal Processing (ICASSP)},
  pages={1--5},
  year={2023},
  organization={IEEE}
}

@article{lee2023design,
  title={Design and single-shot fabrication of lensless cameras with arbitrary point spread functions},
  author={Lee, Kyung Chul and Bae, Junghyun and Baek, Nakkyu and Jung, Jaewoo and Park, Wook and Lee, Seung Ah},
  journal={Optica},
  volume={10},
  number={1},
  pages={72--80},
  year={2023},
  publisher={Optica Publishing Group}
}

@inproceedings{antipa2019video,
  title={Video from stills: Lensless imaging with rolling shutter},
  author={Antipa, Nick and Oare, Patrick and Bostan, Emrah and Ng, Ren and Waller, Laura},
  booktitle={2019 IEEE International Conference on Computational Photography (ICCP)},
  pages={1--8},
  year={2019},
  organization={IEEE}
}

@article{pan2022image,
  title={Image reconstruction with transformer for mask-based lensless imaging},
  author={Pan, Xiuxi and Chen, Xiao and Takeyama, Saori and Yamaguchi, Masahiro},
  journal={Optics Letters},
  volume={47},
  number={7},
  pages={1843--1846},
  year={2022},
  publisher={Optica Publishing Group}
}

@inproceedings{bae2020lensless,
  title={Lensless imaging with an end-to-end deep neural network},
  author={Bae, Donggeon and Jung, Jaewoo and Baek, Nakkyu and Lee, Seung Ah},
  booktitle={2020 IEEE International Conference on Consumer Electronics-Asia (ICCE-Asia)},
  pages={1--5},
  year={2020},
  organization={IEEE}
}

@article{dong2021dwdn,
  title={Dwdn: Deep wiener deconvolution network for non-blind image deblurring},
  author={Dong, Jiangxin and Roth, Stefan and Schiele, Bernt},
  journal={IEEE Transactions on Pattern Analysis and Machine Intelligence},
  volume={44},
  number={12},
  pages={9960--9976},
  year={2021},
  publisher={IEEE}
}

@article{monakhova2020spectral,
  title={Spectral DiffuserCam: lensless snapshot hyperspectral imaging with a spectral filter array},
  author={Monakhova, Kristina and Yanny, Kyrollos and Aggarwal, Neerja and Waller, Laura},
  journal={Optica},
  volume={7},
  number={10},
  pages={1298--1307},
  year={2020},
  publisher={Optical Society of America}
}

@inproceedings{ulyanov2018deep,
  title={Deep image prior},
  author={Ulyanov, Dmitry and Vedaldi, Andrea and Lempitsky, Victor},
  booktitle={Proceedings of the IEEE conference on computer vision and pattern recognition},
  pages={9446--9454},
  year={2018}
}

@article{monakhova2021untrained,
  title={Untrained networks for compressive lensless photography},
  author={Monakhova, Kristina and Tran, Vi and Kuo, Grace and Waller, Laura},
  journal={Optics Express},
  volume={29},
  number={13},
  pages={20913--20929},
  year={2021},
  publisher={Optical Society of America}
}

@article{boyd2011distributed,
  title={Distributed optimization and statistical learning via the alternating direction method of multipliers},
  author={Boyd, Stephen and Parikh, Neal and Chu, Eric and Peleato, Borja and Eckstein, Jonathan and others},
  journal={Foundations and Trends{\textregistered} in Machine learning},
  volume={3},
  number={1},
  pages={1--122},
  year={2011},
  publisher={Now Publishers, Inc.}
}

@book{wiener1964extrapolation,
  title={Extrapolation, interpolation, and smoothing of stationary time series},
  author={Wiener, Norbert},
  year={1964},
  publisher={The MIT press}
}

@article{richardson1972bayesian,
  title={Bayesian-based iterative method of image restoration},
  author={Richardson, William Hadley},
  journal={Journal of the optical society of America},
  volume={62},
  number={1},
  pages={55--59},
  year={1972},
  publisher={Optical Society of America}
}

@article{lucy1974iterative,
  title={An iterative technique for the rectification of observed distributions},
  author={Lucy, Leon B},
  journal={Astronomical Journal},
  volume={79},
  pages={745},
  year={1974}
}

@article{monakhova2019learned,
  title={Learned reconstructions for practical mask-based lensless imaging},
  author={Monakhova, Kristina and Yurtsever, Joshua and Kuo, Grace and Antipa, Nick and Yanny, Kyrollos and Waller, Laura},
  journal={Optics express},
  volume={27},
  number={20},
  pages={28075--28090},
  year={2019},
  publisher={Optical Society of America}
}

@article{yanny2022deep,
  title={Deep learning for fast spatially varying deconvolution},
  author={Yanny, Kyrollos and Monakhova, Kristina and Shuai, Richard W and Waller, Laura},
  journal={Optica},
  volume={9},
  number={1},
  pages={96--99},
  year={2022},
  publisher={Optical Society of America}
}

@article{li2023mwdns,
  title={MWDNs: reconstruction in multi-scale feature spaces for lensless imaging},
  author={Li, Ying and Li, Zhengdai and Chen, Kaiyu and Guo, Youming and Rao, Changhui},
  journal={Optics Express},
  volume={31},
  number={23},
  pages={39088--39101},
  year={2023},
  publisher={Optica Publishing Group}
}

@article{khan2020flatnet,
  title={Flatnet: Towards photorealistic scene reconstruction from lensless measurements},
  author={Khan, Salman Siddique and Sundar, Varun and Boominathan, Vivek and Veeraraghavan, Ashok and Mitra, Kaushik},
  journal={IEEE Transactions on Pattern Analysis and Machine Intelligence},
  volume={44},
  number={4},
  pages={1934--1948},
  year={2020},
  publisher={IEEE}
}

@article{cai2024phocolens,
  title={Phocolens: Photorealistic and consistent reconstruction in lensless imaging},
  author={Cai, Xin and You, Zhiyuan and Zhang, Hailong and Gu, Jinwei and Liu, Wentao and Xue, Tianfan},
  journal={Advances in Neural Information Processing Systems},
  volume={37},
  pages={12219--12242},
  year={2024}
}

@article{baek2022lensless,
  title={Lensless polarization camera for single-shot full-Stokes imaging},
  author={Baek, Nakkyu and Lee, Yujin and Kim, Taeyoung and Jung, Jaewoo and Lee, Seung Ah},
  journal={APL Photonics},
  volume={7},
  number={11},
  year={2022},
  publisher={AIP Publishing}
}

@article{kingshott2022unrolled,
  title={Unrolled primal-dual networks for lensless cameras},
  author={Kingshott, Oliver and Antipa, Nick and Bostan, Emrah and Ak{\c{s}}it, Kaan},
  journal={Optics Express},
  volume={30},
  number={26},
  pages={46324--46335},
  year={2022},
  publisher={Optica Publishing Group}
}

@article{bagadthey2022flatnet3d,
  title={FlatNet3D: intensity and absolute depth from single-shot lensless capture},
  author={Bagadthey, Dhruvjyoti and Prabhu, Sanjana and Khan, Salman S and Fredrick, D Tony and Boominathan, Vivek and Veeraraghavan, Ashok and Mitra, Kaushik},
  journal={Journal of the Optical Society of America A},
  volume={39},
  number={10},
  pages={1903--1912},
  year={2022},
  publisher={Optica Publishing Group}
}

@inproceedings{poudel2024deeplir,
  title={Deeplir: Attention-based approach for mask-based lensless image reconstruction},
  author={Poudel, Arpan and Nakarmi, Ukash},
  booktitle={Proceedings of the IEEE/CVF Winter Conference on Applications of Computer Vision},
  pages={431--439},
  year={2024}
}


\end{document}


\title{Supplementary Material for ``Integrated Forward--Inverse Network for Lensless Image Reconstruction''}

\titlerunning{Integrated Forward--Inverse Network}
\authorrunning{D.~Bae et al.}

\author{}
\institute{}
\maketitle

\renewcommand{\thesection}{S\arabic{section}}
\renewcommand{\thefigure}{S\arabic{figure}}
\renewcommand{\thetable}{S\arabic{table}}
\renewcommand{\theequation}{S\arabic{equation}}
\setcounter{figure}{0}
\setcounter{table}{0}
\setcounter{equation}{0}

\section{Training Details}
\label{app:Training}
Throughout, FSO, ISO, IFIB, and ROI follow the definitions in the main paper.

We train IFIN using the AdamW optimizer with a learning rate of $1\times10^{-4}$ 
and $(\beta_1, \beta_2) = (0.9, 0.999)$ for all parameters except the point spread function (PSF). 
For the PSF, we employ a separate AdamW optimizer with a learning rate of $1\times10^{-3}$. 
Both learning rates are reduced by a factor of $0.5$ when the validation loss plateaus. 
The training is conducted with a batch size of $4$.

For spatially varying deconvolution, we initialize the network by constructing ROI maps aligned to the input size, enabling local adaptation without excessive computational overhead. In the forward--inverse integration, all forward- and inverse-stream gates are initialized to $\alpha{=}0.8$ and $\beta{=}0.2$, balancing the contributions of the forward and inverse operators at the start of training.

For the PSF representation, to reduce computational cost we crop the PSF to retain only its effective region. Under shift-variant conditions, we do not rely on calibrated PSFs; instead, the on-axis PSF is either replicated $k$ times or randomly initialized. We experiment with $k\in\{1,4,9,16\}$ on DiffuserCam, and set $k{=}9$ for both WiderCam and MultiWienerNet (MWNet).

To ensure fairness, we match hyperparameters to those used in prior works. Whenever a network requires PSF inputs, we apply $\ell_2$, $\ell_1$, or max normalization as appropriate. Dataset-specific preprocessing steps such as cropping are not incorporated into the loss and are applied only for visualization. For WiderCam, the affine transform is estimated offline and applied to the label images to define aligned supervision targets during training; it is not optimized jointly with the reconstruction loss. Further details on dataset preparation and affine transforms are provided in \cref{app:dataset} and \cref{app:setup}.

Training IFIN on DiffuserCam requires 144 hours for $k{=}16$ using two NVIDIA RTX A6000 GPUs.
As $k$ decreases, the training time scales down to 112 ($k{=}9$), 92 ($k{=}4$), and 74 hours ($k{=}1$).

\subsubsection{Loss Functions.}
\label{loss}
We minimize a composite objective balancing pixel fidelity, perceptual quality, and a physics prior on the PSF.
Let $x$ be the ground truth, $\hat{x}$ the final reconstructed output, and $h$ the learned PSF. 

The image-domain loss $\mathcal{L}_{\text{img}}$ enforces pixel-wise fidelity between the reconstruction $\hat{x}$ and the ground truth $x$. It anchors training to a numerically accurate solution and prevents the network from drifting under purely perceptual or physics-driven objectives.

The perceptual loss $\mathcal{L}_{\text{perc}}$ is implemented as the VGG-based LPIPS distance~\cite{zhang2018unreasonable} between $\hat{x}$ and $x$. By comparing images in a deep feature space rather than pixel space, it encourages reconstructions that preserve semantic and textural similarity, helping to recover sharper, more visually plausible details than $\ell_2$ loss alone.

The PSF loss $\mathcal{L}_{\text{psf}}$ enforces non-negativity of the learned PSF field, consistent with the interpretation of the PSF as an intensity impulse response. This physical constraint regularizes blind PSF field learning and helps stabilize the deconvolution, reducing noise amplification and ringing artifacts.

\begin{align}
\mathcal{L}_{\text{img}}   &= \|\hat{x}-x\|_2^2
&& \text{fidelity loss} \label{eq:L_img}\\
\mathcal{L}_{\text{perc}}  &= \operatorname{LPIPS}_{\text{VGG}}(\hat{x},x)
&& \text{perceptual loss~\cite{zhang2018unreasonable}} \label{eq:L_perc}\\
\mathcal{L}_{\text{psf}}   &= \|\min(h,0)\|_1
&& \text{PSF non-negativity} \label{eq:L_psf}
\end{align}

\begin{align}\label{eq:total_loss}
\mathcal{L} =
\lambda_{\text{img}}\mathcal{L}_{\text{img}}
+ \lambda_{\text{perc}}\mathcal{L}_{\text{perc}}
+ \lambda_{\text{psf}}\mathcal{L}_{\text{psf}}
\end{align}

\begin{align}\label{eq:lambdas}
(\lambda_{\text{img}},\,\lambda_{\text{perc}},\,\lambda_{\text{psf}})
= (1.0,\,0.05,\,0.1).
\end{align}

Coefficients are selected by validation and kept fixed across all experiments. For the learning-based baselines retrained in our framework, we use the same image and perceptual losses ($\mathcal{L}_{\text{img}}$ and $\mathcal{L}_{\text{perc}}$) and keep the loss weights fixed across datasets, while otherwise following standard method-specific optimization settings from the original implementations when required.

\section{Baselines}
\label{benchmark}
We summarize each baseline used in our study, together with its original reference and a brief characterization.

\subsubsection{Alternating Direction Method of Multipliers (ADMM)}~\cite{boyd2011distributed,monakhova2019learned}:
Following the DiffuserCam-style reconstruction formulation, we solve a total-variation (TV)-regularized inverse problem:
\begin{equation}
\hat{x}
=
\arg\min_{x}
\frac{1}{2}\|\mathbf{A}x-y\|_{2}^{2}
+
\lambda_{\mathrm{TV}}\|\nabla x\|_{1}
\label{eq:admm_tv_obj}
\end{equation}
where $\mathbf{A}$ denotes the calibrated forward operator and $\nabla$ is the finite-difference operator used for total variation. We optimize this objective with ADMM using variable splitting for the data-consistency and TV terms, as in prior DiffuserCam reconstruction pipelines. This baseline is physics-faithful and requires only the calibrated PSF, but it typically needs careful parameter tuning and many iterations to converge.

\subsubsection{Wiener Deconvolution}~\cite{wiener1964extrapolation}:
For a shift-invariant forward model
\begin{equation}
y = h \ast x + \eta,
\end{equation}
where $\ast$ denotes spatial convolution and $\eta$ additive noise,
Wiener deconvolution estimates the latent image in the Fourier domain as
\begin{equation}
\hat{X}(\omega)
=
\frac{H^{*}(\omega)}{|H(\omega)|^{2} + K}\,Y(\omega),
\label{eq:wiener}
\end{equation}
where $H(\omega)$ and $Y(\omega)$ denote the Fourier transforms of the PSF and the measurement, $H^{*}(\omega)$ is the complex conjugate of $H(\omega)$, and $K$ is a scalar noise-to-signal regularization parameter. In practice, we tune $K$ on the validation split for each dataset. It is fast and has a closed form, though noise, PSF mismatch, and strong shift variance limit its accuracy.

\subsubsection{U\mbox{-}Net}~\cite{ronneberger2015u}:
A standard encoder--decoder convolutional neural network (CNN) with skip connections that directly maps the raw lensless measurement to the target image. It serves as a purely data-driven baseline without any explicit use of the forward model.

\subsubsection{NAFNet}~\cite{chen2022simple}:
A modern, parameter-efficient CNN restorer built from Nonlinear Activation Free (NAF) blocks, which replace explicit nonlinear activations with lightweight gating and normalization. Like U-Net, it is purely data-driven but with substantially higher capacity.

\subsubsection{Learned\mbox{-}ADMM\mbox{-}U (Le\mbox{-}ADMM\mbox{-}U)}~\cite{monakhova2019learned}:
An unrolled hybrid baseline derived from the ADMM reconstruction used for DiffuserCam. Each ADMM iteration is interpreted as a network layer with learnable penalty and shrinkage parameters, and the output of the unrolled solver is further refined by a U-Net denoiser. In the original formulation, five ADMM iterations are unrolled, providing a bounded-compute reconstruction network that retains explicit knowledge of the measurement model while using the final CNN to suppress model-mismatch artifacts and improve perceptual quality.

\subsubsection{DeepLIR}~\cite{poudel2024deeplir}:
A hybrid reconstruction model that combines five unrolled ADMM iterations with an attention-based U-Net denoiser. The denoiser is built around ConvNeXt-style blocks and linear attention, and the original implementation uses four downsampling stages, one bottleneck, and three upsampling stages followed by a final ConvNeXt$+$convolution head. Relative to Le-ADMM-U, DeepLIR strengthens the learned post-processing component with a more expressive attention-based backbone to better capture long-range dependencies induced by the multiplexed measurement.

\subsubsection{MultiWienerNet (MWNet)}~\cite{yanny2022deep}:
A spatially varying hybrid baseline designed for systems with multiple field-dependent PSFs. Its pipeline consists of (i) a learnable multi-Wiener deconvolution layer initialized from the measured spatially varying PSFs, which outputs several intermediate deconvolved images, and (ii) a U-Net refinement stage that combines and refines these intermediate reconstructions into a single output. This design explicitly injects region-wise PSF information while remaining substantially faster than iterative spatially varying deconvolution. When full calibration is unavailable in our setting, we initialize the filters using the on-axis PSF and instantiate $k$ filters accordingly.

\subsubsection{Unrolled Primal\mbox{--}Dual Network (UPDN)}~\cite{kingshott2022unrolled}:
A learned primal--dual reconstruction architecture in which the image and measurement variables are updated jointly across a fixed number of iterations. Unlike methods that rely on a single fixed calibrated convolution, UPDN learns the forward and adjoint operators inside the unrolled optimization loop to better compensate for model error. Its primal and dual updates are parameterized by compact CNN modules, yielding a relatively lightweight physics-aware baseline with stronger robustness to mismatch than plain unrolled ADMM. After the unrolled primal--dual iterations, the resulting reconstruction is further processed by a final U-Net denoiser.

\subsubsection{Multi-Wiener Deconvolution Networks (MWDNs)}~\cite{li2023mwdns}:
A feature-space deconvolution baseline that performs Wiener deconvolution inside a multi-scale encoder--decoder rather than only at the input or output image plane. Concretely, the measurement is processed by a common U-Net, and at each skip connection the corresponding feature map is deconvolved with a scale-matched PSF before fusion. This architecture injects optical priors throughout the hierarchy and improves reconstruction fidelity under severe blur by combining multi-scale feature extraction with explicit Wiener-style inversion.

\subsubsection{LensNet}~\cite{bai2025lensnet}:
A stronger embedded-inversion baseline that can be viewed as an extension of MWDNs with a more expressive reconstruction backbone and improved robustness to PSF mismatch. LensNet introduces a learnable Coded Mask Simulator (CMS), which dynamically estimates the empirical PSF behavior and therefore reduces the dependence on a fixed or sparsely calibrated PSF. It embeds Wiener filtering together with CMS-based PSF modeling.

\subsubsection{Modular Learned Reconstruction (MoDL)}~\cite{bezzam2025towards}:
A modular physics-based reconstruction framework composed of three stages: a pre-processor, a camera-inversion module, and a post-processor. The pre-processor suppresses measurement noise before inversion, the middle block performs physics-based inversion (e.g., unrolled ADMM or trainable Wiener-type inversion), and the post-processor removes artifacts and enhances the final image. This modular separation is attractive under model mismatch because it decouples noise suppression before inversion from artifact cleanup after inversion, while still allowing end-to-end training.

\section{Detailed Description of Public Datasets}
\label{app:dataset}
\subsubsection{DiffuserCam.}
The DiffuserCam dataset~\cite{monakhova2019learned} comprises 25{,}000 paired captures acquired simultaneously with a mask-based lensless camera~\cite{antipa2017diffusercam} and a reference lensed camera aligned via a beam splitter, using images from MIRFlickr~\cite{huiskes2008mir} displayed on a computer monitor. The prototype uses an off-the-shelf Light Shaping Diffuser (Luminit \(0.5^{\circ}\)) with a laser-cut paper aperture, positioned approximately \(9\,\mathrm{mm}\) in front of the sensor plane. Raw sensor frames of \(1920 \times 1080\) pixels are downsampled by a factor of four to \(480 \times 270\). For visualization, DiffuserCam images are additionally cropped to \(380 \times 210\), while all reported metrics (PSNR, SSIM, LPIPS) are computed on the non-cropped scenes. The dataset is split into 24{,}000 training and 1{,}000 test images. A single PSF is calibrated at the field center using an on-axis LED point source at the screen plane.

\subsubsection{MultiWienerNet.}
Built on microscope data from Miniscope3D~\cite{yanny2020miniscope3d}, the MWNet dataset~\cite{yanny2022deep} is built from explicitly calibrated spatially varying PSFs across the field: a sub-resolution bead is scanned to measure the PSF at multiple sensor locations, effectively sampling a \(3 \times 3\) grid over the imaging field. Using these measured PSFs, a synthetic training set is generated by convolving natural images with the spatially varying forward model and adding Poisson and Gaussian noise to emulate realistic measurements. This yields 22{,}125 2D paired samples, split 80/20 for training and testing. All training data are simulated at the system's sensor field-of-view resolution, and the trained models are finally evaluated on real lensless measurements from the calibrated setup. In our experiments, we resize all images to \(320 \times 224\) to reduce computational cost.

\section{Detailed Description of WiderCam Dataset}
\label{app:setup}
\subsubsection{WiderCam.}
\begin{figure}[tb]
  \centering
  \includegraphics[width=1\linewidth]{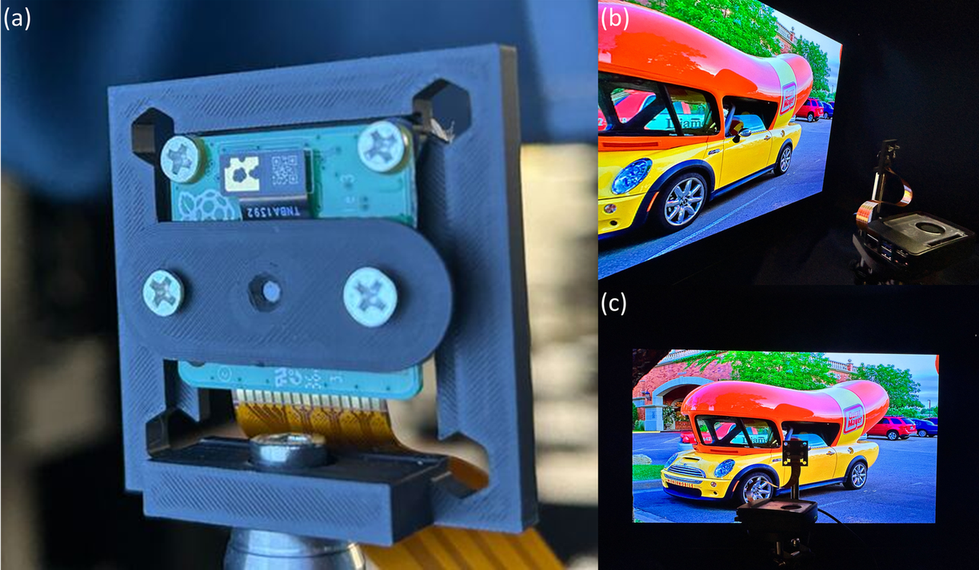}
\caption{Prototype lensless camera and dataset capture setup.
(a) Lensless camera prototype with a CMOS sensor mounted in a custom 3D-printed holder.
(b--c) Display-capture configuration used for the WiderCam dataset: reference images are rendered on a calibrated display while the prototype records the corresponding lensless measurements at a fixed geometry.}
  \label{fig:camera_setup}
\end{figure}

We introduce a new dataset captured with a compact high-resolution lensless camera built around a custom phase-mask system. The optical design prioritizes on-axis resolution at the cost of increased PSF shift variance with incident angle. Using a Sony IMX708 sensor, we capture lensless measurements while displaying MIRFlickr images such that they occupy 80\% of a 48-inch OLED TV screen at a working distance of $30\,\mathrm{cm}$, as shown in \cref{fig:camera_setup}. A total of 25{,}000 images are recorded and then split into 24{,}000 for training and 1{,}000 for testing, following the same indices as in the DiffuserCam dataset. We capture raw measurements at \(4608 \times 2592\) resolution and resize them to \(480 \times 270\).

\begin{figure}[tb]
  \centering
  \includegraphics[width=1\linewidth]{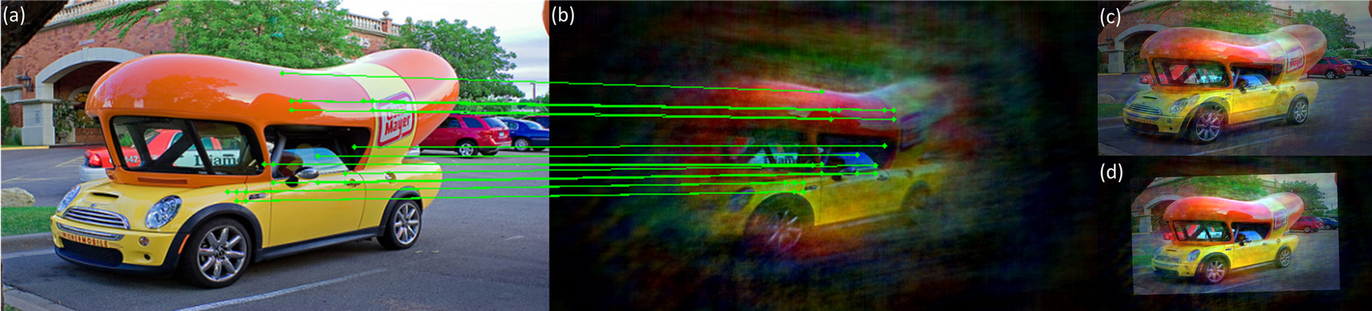}
\caption{Affine registration for label and display-capture pairs.
(a) Reference image shown on the display. (b) Raw lensless measurement captured by the prototype; green segments indicate LoFTR-based feature correspondences used to estimate the affine transform~\cite{sun2021loftr}. (c) Overlay before registration. (d) Overlay after applying the estimated affine warp, yielding pixel-wise alignment suitable for supervised training and evaluation.}
  \label{fig:camera_affine}
\end{figure}

For target alignment, we first reconstruct an image using a deconvolution baseline, estimate an affine transform between the reconstruction and its corresponding label, and apply this precomputed transform to the label image to define the aligned supervision target used during training, as illustrated in \cref{fig:camera_affine}. Unlike DiffuserCam, where a separate reference camera and spatial cropping are used for alignment, our setup relies on an offline affine warp estimated from the display-capture pair. At test time, the reconstructed output is inverse-warped back to the original label coordinate system for evaluation and visualization.

\subsubsection{Optical Design, Fabrication, and Camera Assembly.}
We designed the phase mask under the mechanical and optical constraints of the system. These constraints included: (i) the mechanical stack and housing of the Sony IMX708 sensor module, (ii) the target field-of-view (FoV) and equivalent focal length, and (iii) the maximum fabricable optical thickness. Within these constraints, we derived a single planoconvex micro-lens profile with a vertex height of 20\,\(\mu\)m and a radius of curvature of 860\,\(\mu\)m that maximizes the effective numerical aperture (NA).

The unit profile was then randomly tiled over a 3.5~mm~\(\times\)~3.5~mm area to form the phase mask. A minimum inter-lens spacing was enforced during placement to preserve fill factor and to suppress degradation of the effective NA caused by mutual overlap and edge clipping.

The mask was fabricated as a multi-level phase element using the grayscale lithography process described in~\cite{lee2023design}. The continuous height map of the planoconvex profile was converted to grayscale dose, enabling a single exposure--development process. Post-fabrication inspection using surface profilometry and microscopy verified the profile integrity and confirmed the absence of large-area defects.

The fabricated mask was then aligned and attached in front of the IMX708 sensor to form a lensless camera. A mechanical aperture of 2~mm diameter was applied directly at the mask plane to define the active pupil and to mitigate stray light and edge effects during imaging.

\begin{figure}[tb]
  \centering
  \includegraphics[width=1\linewidth]{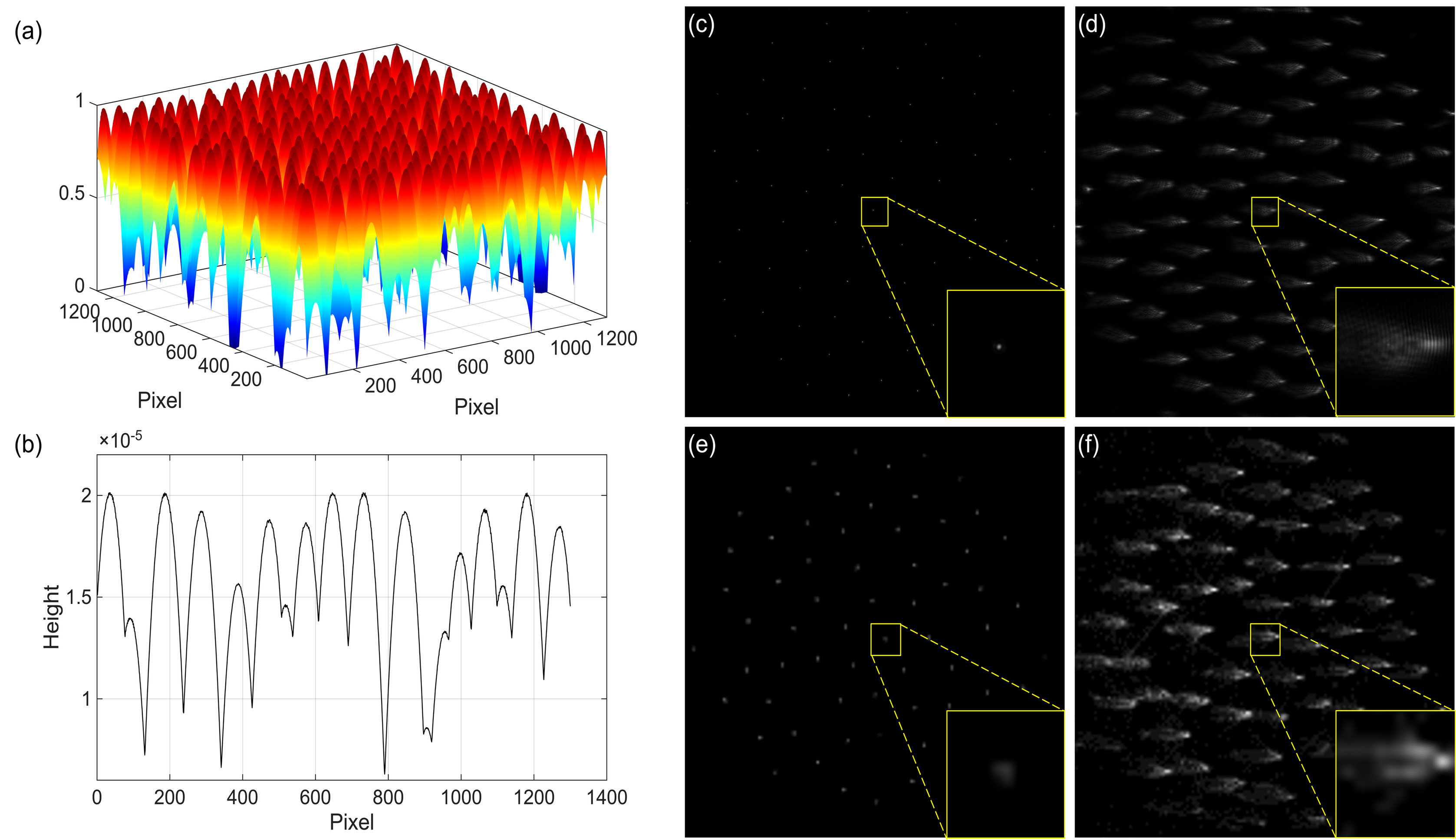}
  \caption{(a) The designed phase mask pattern optimized for high-resolution lensless imaging. 
  (b) A representative line profile.
  (c) Simulated on-axis PSF.
  (d) Simulated off-axis PSF at $40^\circ$, showing the effect of shift variance across the field.
  (e) Captured on-axis PSF.
  (f) Captured off-axis PSF.}
  \label{fig:mask}
\end{figure}

\subsubsection{PSF Measurement and Optical Performance.}
We characterized the system by measuring the PSFs on- and off-axis. A comparison between the designed mask, simulated PSFs, and experimentally captured PSFs is shown in \cref{fig:mask}, highlighting the agreement between design and physical performance.
\begin{itemize}
  \item \textbf{On-axis.} The measured PSF closely matched the designed PSF. Cross-correlation analysis with the design yielded high similarity, and the measured full width at half maximum (FWHM) was within the expected range, consistent with the high effective NA. These results provide an empirical bound on the on-axis optical resolution.
  \item \textbf{Off-axis.} At larger field angles, aberrations became pronounced as expected under high-NA operation. The PSF exhibited asymmetric tails consistent with coma aberration from the planoconvex element. Consequently, the system exhibits shift-variant imaging behavior: high optical performance on-axis, with aberration-limited quality off-axis.
\end{itemize}

\subsubsection{Implications.}
\begin{figure}[tb]
  \centering
  \includegraphics[width=\linewidth]{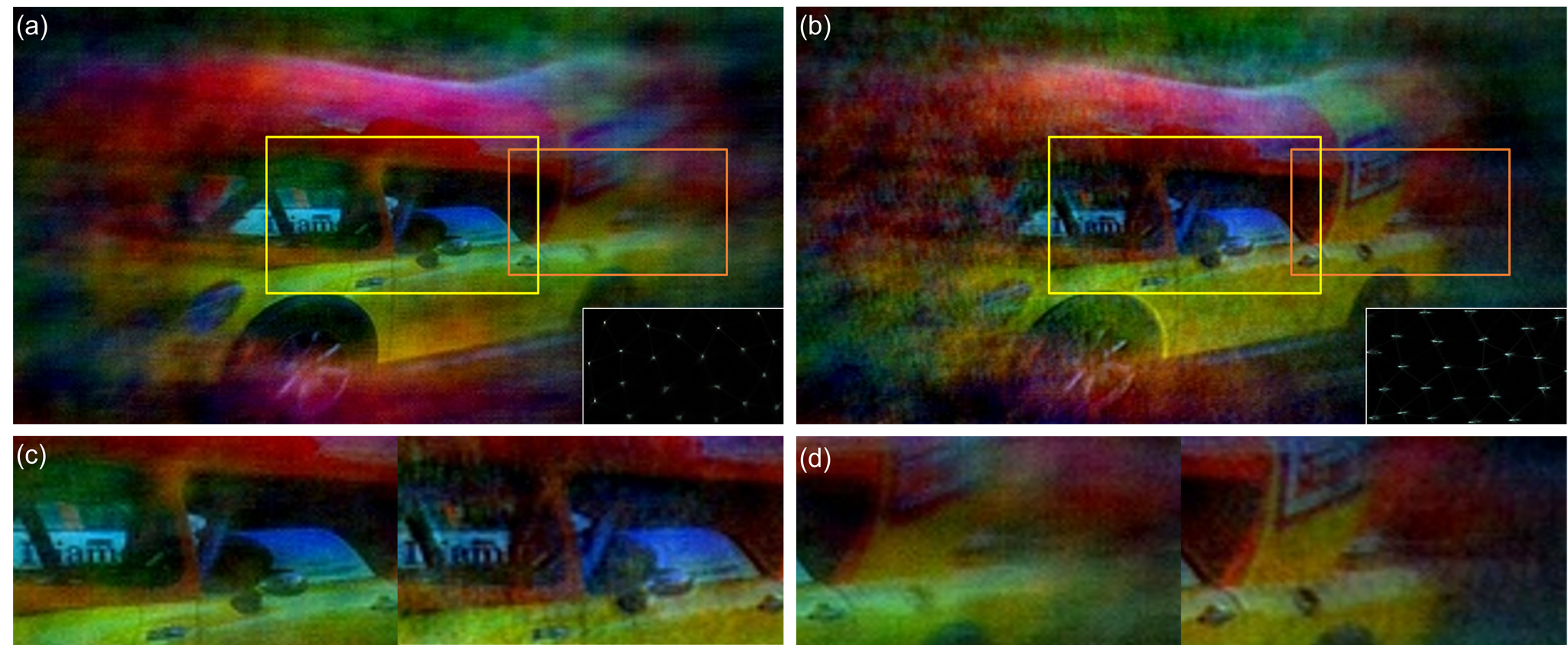}
   \caption{(a) Reconstruction using the center PSF along with the corresponding center PSF.  
    (b) Reconstruction using an off-axis PSF along with the corresponding off-axis PSF.  
    (c) Comparison of sharpness at the image center between (a) and (b).  
    (d) Comparison at the image periphery, demonstrating differences caused by field-dependent PSFs.}
  \label{fig:variant-deconv}
\end{figure}

Maximizing NA under a limited thickness budget (20\,\(\mu\)m) was effective for on-axis resolution but increased sensitivity to off-axis aberrations and FoV non-uniformity. Hardware routes to mitigate this include aspheric refinements, multi-layer phase designs, and orientation-aware cell geometries; software routes include deconvolution with a field-dependent PSF or learned reconstructions that explicitly model shift variance. As illustrated in \cref{fig:variant-deconv}, deconvolving with PSFs drawn from different field regions produces noticeable differences in focal sharpness (and associated artifacts), directly evidencing the field dependence.

\section{Detailed Description of Additional Experiments}
To demonstrate the generalization capability of the proposed IFIN, we evaluate its performance on two classical yet challenging inverse problems: Gaussian deblurring and inline holography.

\subsubsection{Gaussian Deblurring.}
Gaussian deblurring is a fundamental ill-posed problem characterized by a space-invariant convolution kernel. For this experiment, we construct synthetic blurred measurements from clean MIRFlickr images resized to \(384\times384\). Specifically, each clean target image is degraded using Gaussian blur operators with increasing standard deviations \(\sigma \in \{5,10,15\}\), followed by additive Gaussian noise with standard deviation \(0.03\). Thus, the forward degradation model is
\begin{equation}
y = H_{\sigma} \ast x + \eta,
\end{equation}
where \(\ast\) denotes spatial convolution, \(H_{\sigma}\) denotes the Gaussian blur operator corresponding to blur level \(\sigma\), and \(\eta\) denotes additive Gaussian noise. By varying \(\sigma\), we evaluate IFIN under progressively more severe blur. The degraded measurements are synthesized on the fly from the clean targets during both training and testing, so the forward degradation process is fully controlled. Within the IFIN framework, the differentiable forward projection layer explicitly models the blur operator \(H_{\sigma}\). The subsequent learnable inverse update stages are trained to suppress reconstruction artifacts such as ringing.

\subsubsection{Inline Holography.}
\begin{figure}[tb]
\centering
\includegraphics[width=\linewidth]{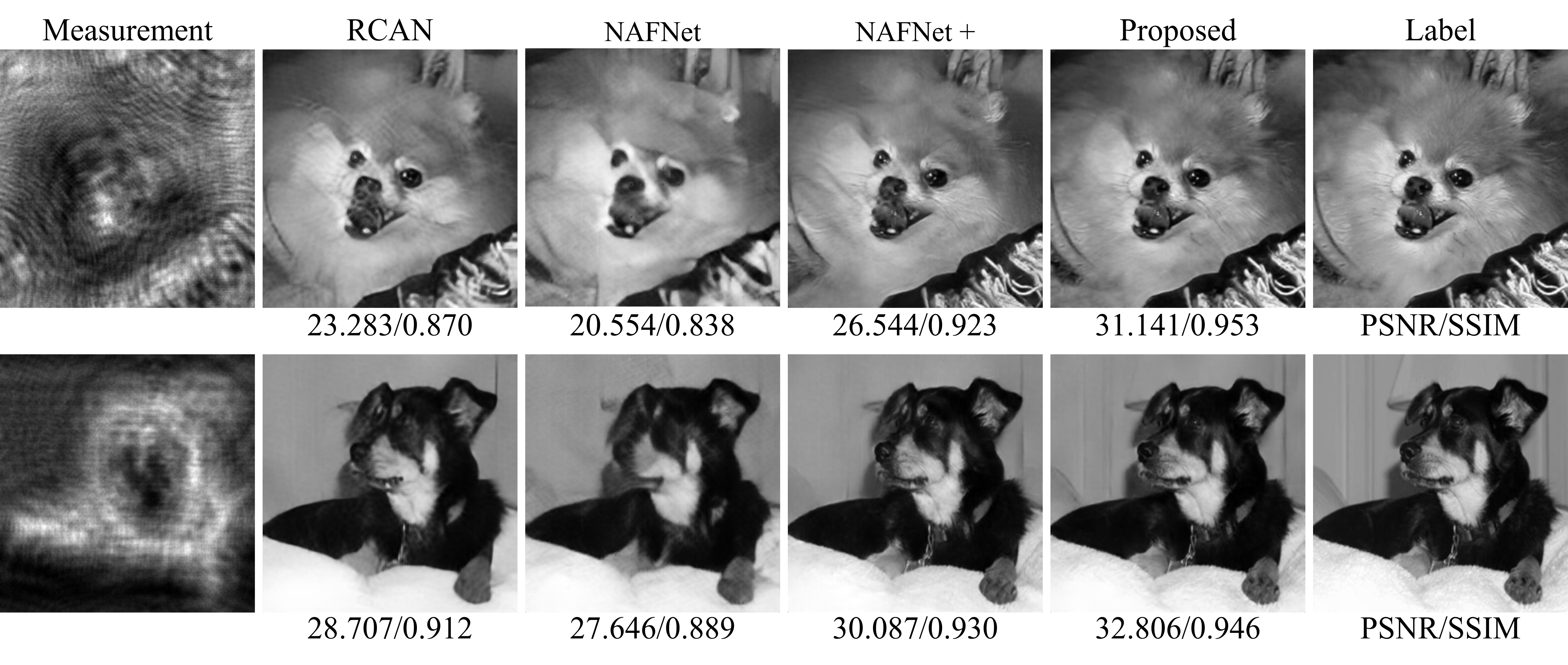}
\caption{
Reconstruction results for the inline holography benchmark. 
From left to right: input intensity hologram, reconstructed object amplitude by each method, and ground-truth object amplitude. 
The proposed IFIN successfully adapts to the holographic forward model and recovers cleaner object structures from a single hologram.}
\label{fig:supp_holo}
\end{figure}

Inline holography reconstructs an object from intensity-only measurements through a diffraction-based propagation model. In this experiment, we adapt IFIN to an inline holography forward model in order to evaluate whether the proposed forward--inverse integration can be extended beyond the lensless imaging systems considered in the main experiments. Reconstruction results are shown in \cref{fig:supp_holo}.

For this benchmark, we generate synthetic holograms from the Dogs vs.\ Cats dataset using a coherent forward model. Each source image is resized to $256\times256$ and converted to grayscale. The normalized image $x_{\mathrm{holo}}[i,j]$ is interpreted as a normalized object absorbance (density) map, and the object field is defined as a purely amplitude-modulating field

\begin{equation}
u_{\mathrm{obj}}[i,j]
\;=\;
1-\alpha_{\mathrm{abs}}\,x_{\mathrm{holo}}[i,j],
\end{equation}
where $\alpha_{\mathrm{abs}}=0.8$ is the absorption factor.

The field is then propagated to the sensor plane using the angular spectrum method (a Fourier-domain transfer function) with wavelength $\lambda=532\,\mathrm{nm}$, pixel pitch $\Delta=5.08\,\mu\mathrm{m}$, and propagation distance $z=30\,\mathrm{mm}$:
\begin{equation}
u_{\mathrm{sens}}
\;=\;
\mathcal{H}_{z}\,u_{\mathrm{obj}},
\end{equation}
where $\mathcal{H}_{z}$ denotes the holographic propagation operator. The measured hologram is formed as the sensor-plane intensity
\begin{equation}
y_{\mathrm{holo}}[i,j]
\;=\;
|u_{\mathrm{sens}}[i,j]|^2.
\end{equation}
After normalization, the hologram is corrupted with Poisson shot noise and additive Gaussian read noise:
\begin{equation}
y_{\mathrm{holo},n}[i,j]
\sim
\frac{\mathrm{Poisson}(y_{\mathrm{holo}}[i,j]\,N_{\mathrm{ph}})}{N_{\mathrm{ph}}}
+\mathcal{N}(0,\sigma_{\mathrm{read}}^2),
\end{equation}
where the photon count is set to $N_{\mathrm{ph}}=2{,}000$ and the read-noise standard deviation is $\sigma_{\mathrm{read}}=0.005$. The final network input is the noisy intensity hologram, while the supervision target is the corresponding clean object amplitude image.

To adapt IFIN, we replace the original PSF-conditioned physical modules with a PSF-free symmetric propagation pair. In the lensless-imaging model, the ISO is implemented as a Wiener-style deconvolution module conditioned on the PSF field, and the FSO is implemented as an FFT-based convolution module using the corresponding PSF. The original architecture therefore relies on learned or calibrated PSFs, multi-scale PSF embeddings, ROI-based blending, and PSF-specific regularization. In the holography setting we remove all of these components and replace the PSF-conditioned inverse and forward operators with diffraction-based ones:
\begin{equation}
\operatorname{ISO}_{\mathrm{holo}}(x)
=
\left|
\mathcal{F}^{-1}\!\big(
\mathcal{F}(x)\,\mathcal{H}_{z_I}^{*}
\big)
\right|^2,
\end{equation}
\begin{equation}
\operatorname{FSO}_{\mathrm{holo}}(x)
=
\left|
\mathcal{F}^{-1}\!\big(
\mathcal{F}(x)\,\mathcal{H}_{z_F}
\big)
\right|^2,
\end{equation}
where $\mathcal{H}_{z_I}^{*}$ denotes the conjugate propagator for back-propagation, $\mathcal{H}_{z_F}$ denotes the forward propagator, and $z_I$ and $z_F$ are learnable scalar propagation distances.

Accordingly, at each IFIB, the inverse-stream and forward-stream features exchange information through diffraction-based inverse and forward operators. At the $n$-th hierarchy level, the holography-specific operators first produce intermediate outputs,
\begin{equation}
\tilde{y}_{\mathrm{holo}}^{(n)}
=
\operatorname{FSO}_{\mathrm{holo}}\!\left(x_{\mathrm{holo}}^{(n)}\right),
\end{equation}
\begin{equation}
\tilde{x}_{\mathrm{holo}}^{(n)}
=
\operatorname{ISO}_{\mathrm{holo}}\!\left(y_{\mathrm{holo}}^{(n)}\right),
\end{equation}
which are then integrated through the same bidirectional update rule as in the main paper, where the forward- and inverse-stream gates are tied to a single learnable pair $\alpha^{(n)},\beta^{(n)}$. Therefore, the surrounding two-stream forward--inverse integration remains unchanged, while the underlying physical operators are replaced by a diffraction model appropriate for inline holography.

\section{Ablation on DiffuserCam}
\label{app:ablation}
\Cref{tab:ablation_components} summarizes ablations of the proposed IFIN model on DiffuserCam with $k{=}1$.
\begin{table}[tb]
\caption{
Unified component ablation on DiffuserCam ($k{=}1$).
The table summarizes ablations on (i) forward--inverse integration, (ii) internal operator design, and (iii) network components including the refinement block and PSF encoder.
All variants are obtained by modifying a single component from the full IFIN model.
}
\label{tab:ablation_components}
\begin{center}
\begin{adjustbox}{max width=\linewidth}
\begin{tabular}{llccc}
\toprule
\bf Category & \bf Setting
    & \bf PSNR~$\uparrow$
    & \bf LPIPS~$\downarrow$
    & \bf SSIM~$\uparrow$ \\
\midrule

\multirow{4}{*}{Forward--inverse integration}
& Identity FSO/ISO         & 24.674 & 0.255 & 0.800 \\
& FSO-only (w/o ISO)        & 27.123 & 0.223 & 0.833 \\
& ISO-only (w/o FSO)        & 28.711 & 0.185 & 0.882 \\
& w/o initial ISO           & 29.443 & \textbf{0.176} & 0.891 \\

\midrule
\multirow{4}{*}{Operator design}
& w/o 2D regularizer       & 28.880 & 0.186 & 0.881 \\
& w/o padding in ISO \& FSO & 28.445 & 0.191 & 0.879 \\
& w/o padding in ISO        & 29.001 & 0.185 & 0.883 \\
& w/o padding in FSO        & 29.037 & 0.184 & 0.885 \\

\midrule
\multirow{2}{*}{Network block}
& ConvG instead of RB       & 29.014 & 0.184 & 0.882 \\
& resize-only PSF input     & 29.014 & 0.182 & 0.886 \\

\midrule
\textbf{Full model} & \textbf{IFIN ($k{=}1$)}
& \textbf{29.535} & 0.179 & \textbf{0.892} \\

\bottomrule
\end{tabular}
\end{adjustbox}
\end{center}
\end{table}

\subsubsection{Effect of FSO/ISO and their integration.}
The results show that both the explicit forward--inverse integration and the learned PSF-aware feature modules contribute meaningfully to the final reconstruction quality.

\paragraph{FSO/ISO as identity.}
We replace both FSO and ISO with identity mappings and allow only latent mixing between the two streams. This control shows that simple feature mixing is not sufficient: improvements are limited compared to the full IFIN model. Enforcing forward and inverse operators within the feature space is what propagates measurement-domain cues and stabilizes high-frequency reconstruction.

\paragraph{w/o ISO.}
We remove the inverse operators and retain only a forward-guided pathway. Without ISO, information flows primarily from the measurement domain, reducing feature sharpening and attenuating high-frequency components. This leads to softer textures and lower reconstruction fidelity, as reflected by the drop in PSNR and SSIM and the increase in LPIPS.

\paragraph{w/o FSO.}
We remove the forward operators and retain only inverse guidance, where the measurement is injected only at the first stage of the network. Without FSO to enforce explicit measurement consistency throughout the hierarchy, the model shows a noticeable drop in fidelity and less stable feature propagation, although it still benefits from the learned inverse pathway.

\paragraph{w/o initial ISO.}
We remove the initial ISO that produces the coarse reconstruction fed alongside the measurement into the encoder, initializing the image stream from the raw measurement instead. This yields only a small change relative to the full model (a slight PSNR/SSIM drop, with comparable or better LPIPS); the initial ISO is not the primary source of the gain but mainly stabilizes the initial feature state and slightly accelerates convergence, while the bidirectional FSO/ISO coupling within the hierarchy drives the improvement.

\subsubsection{Effect of internal components in FSO and ISO.}
We next ablate internal design choices in FSO and ISO, including the learnable 2D regularization term and the padding strategies used to mitigate boundary artifacts (\cref{tab:ablation_components}).

\paragraph{w/o 2D regularizer.}
Removing the learnable 2D regularizer degrades all metrics, indicating that frequency-domain inversion benefits from an adaptive stabilization term rather than a purely analytic inverse. This regularizer helps suppress noise amplification and improves robustness when the forward model is imperfect.

\paragraph{w/o padding in ISO \& FSO.}
When padding is removed from both operators, the performance drops noticeably among the operator-design ablations. This confirms that proper boundary handling is important for avoiding wrap-around and edge artifacts in FFT-based physical modules.

\paragraph{w/o padding in ISO.}
Removing padding only in ISO still degrades the reconstruction; the inverse step is particularly sensitive to boundary artifacts. Since ISO performs deconvolution-like recovery, insufficient padding more easily introduces ringing and boundary distortion.

\paragraph{w/o padding in FSO.}
Removing padding only in FSO also reduces performance, although the degradation is slightly smaller than when padding is removed jointly or in the ISO alone. This indicates that accurate forward-domain feature propagation remains important, even when the inverse pathway is preserved.

\subsubsection{Effect of the refinement block (RB).}
We compare the proposed refinement block (RB) against a simpler convolutional baseline (ConvG), which uses a shallow convolution--gating structure without the normalization-free residual design of RB (\cref{tab:ablation_components}). Replacing RB with the simpler ConvG block leads to a noticeable drop in PSNR and SSIM and an increase in LPIPS, indicating that the stronger prior encoded by RB is beneficial for stabilizing the forward--inverse integration and refining high-frequency details. This result suggests that the gain does not come only from the physical operators themselves, but also from the capacity of the feature-refinement module that follows them.

\subsubsection{Effect of the PSF encoder.}
We also ablate the PSF encoder by replacing it with simple resizing of the calibrated PSF without any CNN processing (\cref{tab:ablation_components}). The CNN-based encoder yields consistently higher PSNR and SSIM and lower LPIPS, so learning multi-scale PSF embeddings from data is preferable to a fixed hand-crafted representation. These embeddings are reused across IFIBs and improve both fidelity and perceptual quality.

\subsubsection{Effect of the number of PSFs.}
We vary the number of learnable PSFs $k{=}s^2$. Small $k$ (e.g., $k{=}1$) underfits spatial variability, while excessively large $k$ increases computation without proportional gains. Because the computational cost of the ISO branch scales linearly with $k$, the choice of $k$ matters most when the PSFs have large spatial footprints. Performance improves consistently from $k{=}1$ to $k{=}9$ across all three datasets (\cref{tab:ablation_psfs}); modeling stronger spatial variation helps. However, these gains come at the cost of increased computational complexity, as discussed later in \cref{tab:complexity_all}.

\begin{table}[tb]
\centering
\caption{
Ablation on the number of learnable PSFs ($k{=}s^2$) across the three benchmarks.
For readability, this table reports results up to $k{=}9$ across all three datasets; the DiffuserCam result for $k{=}16$ is reported in the main paper.
}
\label{tab:ablation_psfs}
\resizebox{\linewidth}{!}{
\begin{tabular}{l
                ccc
                ccc
                ccc}
\toprule
Dataset & \multicolumn{3}{c}{DiffuserCam} & \multicolumn{3}{c}{WiderCam} & \multicolumn{3}{c}{MultiWienerNet} \\
\cmidrule(lr){1-1} \cmidrule(lr){2-4} \cmidrule(lr){5-7} \cmidrule(lr){8-10}
Metrics & PSNR $\uparrow$ & LPIPS $\downarrow$ & SSIM $\uparrow$ 
 & PSNR $\uparrow$ & LPIPS $\downarrow$ & SSIM $\uparrow$
 & PSNR $\uparrow$ & LPIPS $\downarrow$ & SSIM $\uparrow$ \\
\midrule
$k{=}1$   & 29.535 & 0.179 & 0.892 & 24.963 & 0.211 & 0.815 & 29.733 & 0.197 & 0.848 \\
$k{=}4$   & 29.612 & 0.177 & 0.893 & 25.148 & 0.205 & 0.820 & 30.208 & 0.185 & 0.855 \\
$k{=}9$   & 29.751 & 0.176 & 0.893 & 25.444 & 0.201 & 0.824 & 31.083 & 0.175 & 0.866 \\
\bottomrule
\end{tabular}}
\end{table}

\subsubsection{Effect of the ROI scale $\sigma_x$.}
We study the spatial scale of the Gaussian ROI maps used to blend region-wise PSFs. Here $\sigma_x$ denotes the isotropic ROI Gaussian width $\sigma_r$ from the main paper.
\Cref{tab:ablation_sigma_roi} reports a quantitative ablation where, for each
choice of $\sigma$, we train a separate model from scratch. A small $\sigma$
makes the ROIs overly local and fragmented, whereas a large $\sigma$ spreads each
ROI too broadly, reducing the effective dynamic range of the blending weights and
blurring the result. A moderate value (e.g., $\sigma_x{=}60$ in our
setting for DiffuserCam) yields the best trade-off between fidelity and stability.

\begin{figure}[tb]
  \centering
  \includegraphics[width=\linewidth]{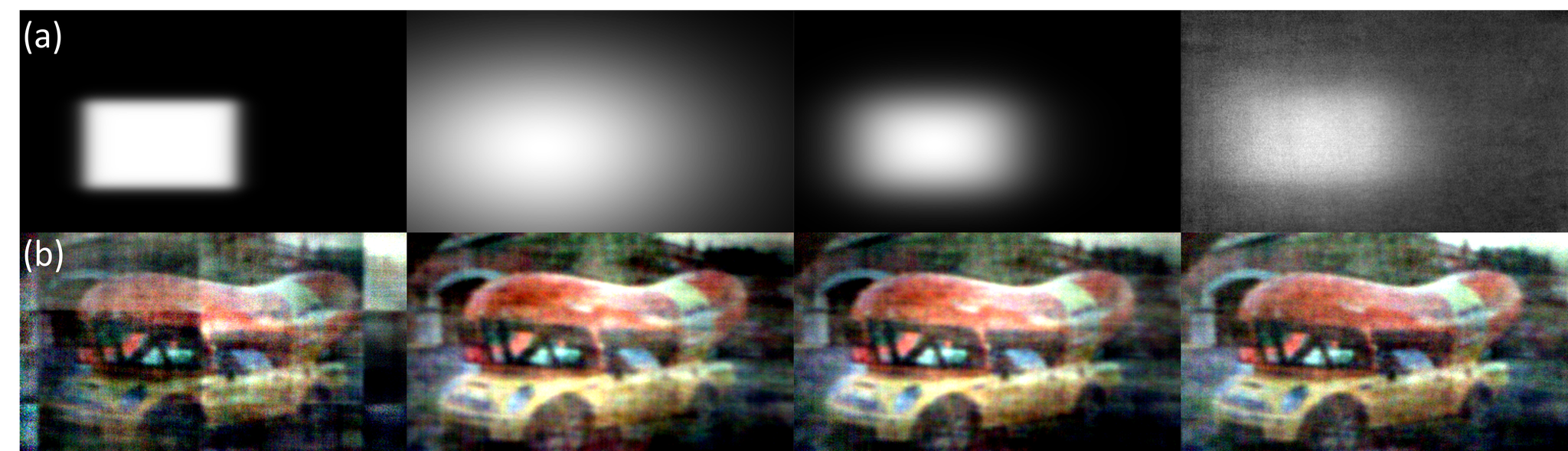}
  \caption{
  Effect of the ROI Gaussian scale on PSFs and reconstructions (DiffuserCam, $k{=}16$). 
  (a) Examples of ROI maps for different settings: small $\sigma$, large $\sigma$, the default fixed $\sigma{=}60$ along the $x$-axis, and the learned ROI maps.
  (b) Corresponding reconstructions using each ROI configuration (columns aligned with (a)). 
  In all cases, we use the same trained ISO and only the ROI scale $\sigma_x$ is modified at test time.
  Small $\sigma$ leads to fragmented regions and visible seams, large $\sigma$ oversmooths details, whereas the default and learned ROIs provide sharper and more spatially consistent results.}
  \label{fig:roi_recon}
\end{figure}

To better visualize the effect of $\sigma_x$ itself, \cref{fig:roi_recon}
shows a complementary experiment in which we freeze a trained ISO (and all other network weights) and vary only $\sigma$ at test time when generating the
Gaussian ROIs. The qualitative trends are consistent with the retrained models:
small $\sigma$ produces block-like artifacts and seams at tile boundaries, very
large $\sigma$ compresses the dynamic range of the ROI weights and slightly
blurs details, while the default setting and the learned ROI maps produce sharper
and more spatially consistent reconstructions.

\begin{table}[tb]
\caption{
Ablation on the ROI Gaussian scale $\sigma_x$ (DiffuserCam, $k{=}16$).
}
\label{tab:ablation_sigma_roi}
\begin{center}
\begin{adjustbox}{max width=\linewidth}
\begin{tabular}{lccc}
\toprule
\bf $\sigma_x$ in ROI ($k{=}16$)
           & \bf PSNR~$\uparrow$
           & \bf LPIPS~$\downarrow$
           & \bf SSIM~$\uparrow$ \\
\midrule
30              & 29.674 & 0.177 & 0.892 \\
60 (Proposed)   & 29.862 & 0.174 & 0.893 \\
120             & 29.729 & 0.177 & 0.893 \\
\bottomrule
\end{tabular}
\end{adjustbox}
\end{center}
\end{table}

\section{Learned PSFs without Calibration}
\label{app:psf_estimation}
\begin{figure}[tb]
  \centering
  \includegraphics[width=0.75\linewidth]{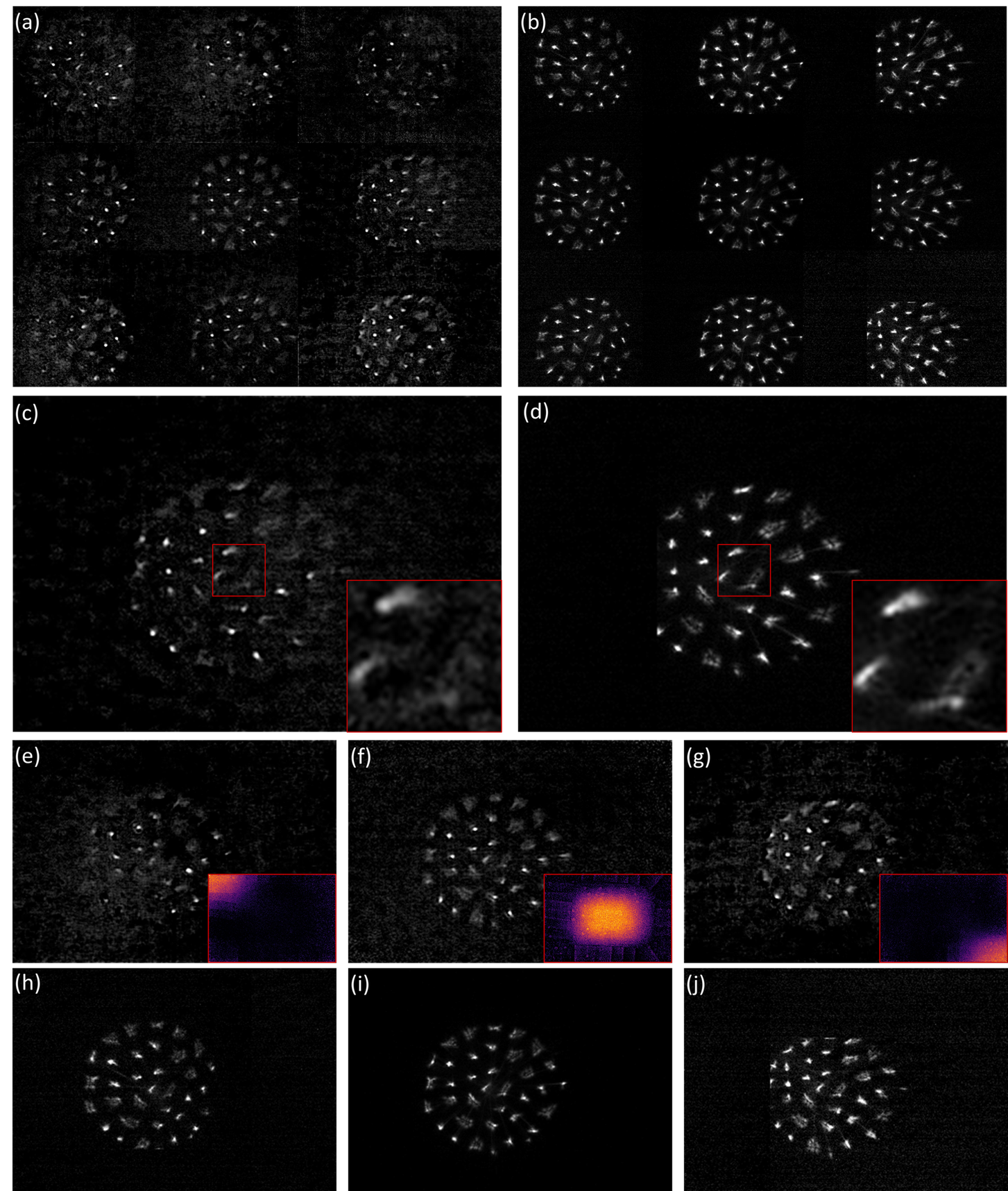}
  \caption{Comparison of learned and calibrated PSFs in the MWNet dataset.
(a) PSFs estimated through training.
(b) Calibrated PSFs.
(c) Estimated PSF at $r=6$ near the right-center position.
(d) Calibrated PSF at the corresponding location.
(e--g) Estimated PSFs at different positions according to the learned ROI weights.
(h--j) Calibrated PSFs corresponding to the same indices as in (e--g).}
  \label{fig:psf_visual}
\end{figure}

\Cref{fig:psf_visual} visualizes the learned $3\times3$ PSF field from the MWNet dataset. Near the optical center, PSFs are compact and approximately isotropic, whereas off-axis locations exhibit increased spread, slight centroid shifts, and mild anisotropy---patterns commonly observed with diffusers and wide-aperture optics. This spatial trend correlates with the improvements seen on real data: FSO reproduces location-dependent blurs using the learned PSFs, and ISO inverts them with data-driven regularization, leading to sharper reconstructions with fewer boundary artifacts.

The learned PSFs remain normalized and vary smoothly across neighbors, reflecting physically plausible optics. Because the PSF field is shared across scales and injected into every IFIB, the network preserves forward--inverse consistency throughout the hierarchy. Qualitatively, these PSFs agree with expected diffuser patterns and reveal off-axis blur variations that standard shift-invariant models fail to capture, explaining IFIN's robustness under strong shift variance.

\paragraph{Learned vs.\ calibrated PSFs.}
To quantify the benefit of end-to-end PSF learning, we compare a learned PSF field against frozen calibrated PSFs on MWNet ($k{=}9$), keeping all other settings fixed (\cref{tab:psf_learned_vs_calib}). The learned field improves all metrics, indicating that data-driven refinement corrects residual calibration mismatch rather than merely reproducing the measured kernels. Together with the smooth, normalized, and physically plausible structure of the learned PSFs (\cref{fig:psf_visual}), this supports learned PSF fields as a practical alternative when dense calibration is unavailable.

We emphasize that, for unseen or sparsely calibrated diffusers, this is \emph{supervised adaptation} rather than fully unsupervised cross-camera transfer. Given paired target-camera data, IFIN can be trained starting from a single measured PSF or a randomly initialized PSF field, substantially reducing the dense per-system calibration burden; however, it still requires target-domain supervision and a reasonably faithful forward model, and we do not claim zero-shot transfer to a new camera without any retraining.

\begin{table}[tb]
\caption{Learned vs.\ frozen calibrated PSFs on MWNet ($k{=}9$), with all other settings fixed (values reproduced from the main paper for completeness).}
\label{tab:psf_learned_vs_calib}
\begin{center}
\begin{adjustbox}{max width=0.7\linewidth}
\begin{tabular}{lccc}
\toprule
\bf Setting & \bf PSNR~$\uparrow$ & \bf LPIPS~$\downarrow$ & \bf SSIM~$\uparrow$ \\
\midrule
Frozen calibrated PSFs & 30.350 & 0.183 & 0.857 \\
\textbf{Learned PSF field}      & \textbf{31.083} & \textbf{0.175} & \textbf{0.866} \\
\bottomrule
\end{tabular}
\end{adjustbox}
\end{center}
\end{table}

\section{Comparison of Deconvolution and ISO}
\begin{figure}[tb]
  \centering
  \includegraphics[width=\linewidth]{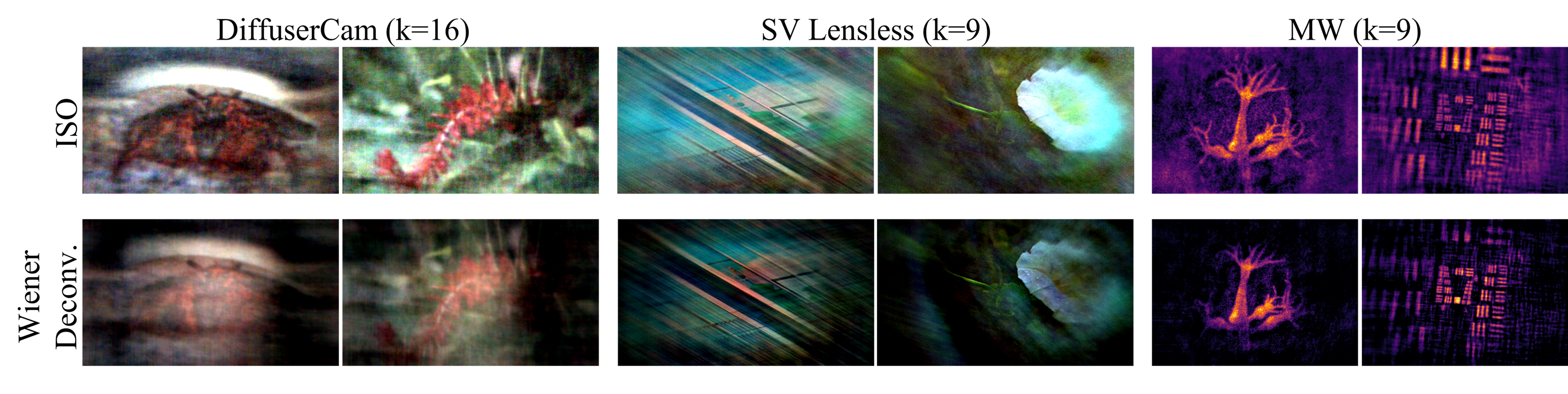}
  \caption{Comparison between the proposed ISO and classical Wiener deconvolution on DiffuserCam, WiderCam, and MWNet datasets. ISO mitigates noise amplification and ringing while improving fine-detail recovery.}
  \label{fig:deconv}
\end{figure}

\Cref{fig:deconv} qualitatively compares classical Wiener deconvolution with our learned ISO. On DiffuserCam, even when the deconvolution hyperparameters are carefully tuned, classical Wiener filtering only partially restores fine structure and tends to leave residual blur and ringing, whereas ISO produces sharper textures and cleaner edges under the same forward model. For the WiderCam dataset, deconvolution is further limited by strong shift variance from the wide-FoV design and angular-response effects that cause peripheral light loss; in this regime, ISO better compensates for these spatially varying degradations and yields more uniform reconstructions across the field. On the MWNet dataset, ISO recovers a wider range of simulated structures and resolves tighter USAF target patterns than deconvolution, illustrating the benefit of incorporating system-aware operators into the network.

Beyond serving as a module inside IFIN, the proposed ISO can also be used as a standalone pretrained inverse mapping, providing a fast, learned alternative to hand-tuned deconvolution for direct inference.

\section{Modeling Shift Variance in FSO}
\label{SVFSO}
\begin{figure}[tb]
  \centering
  \includegraphics[width=1\linewidth]{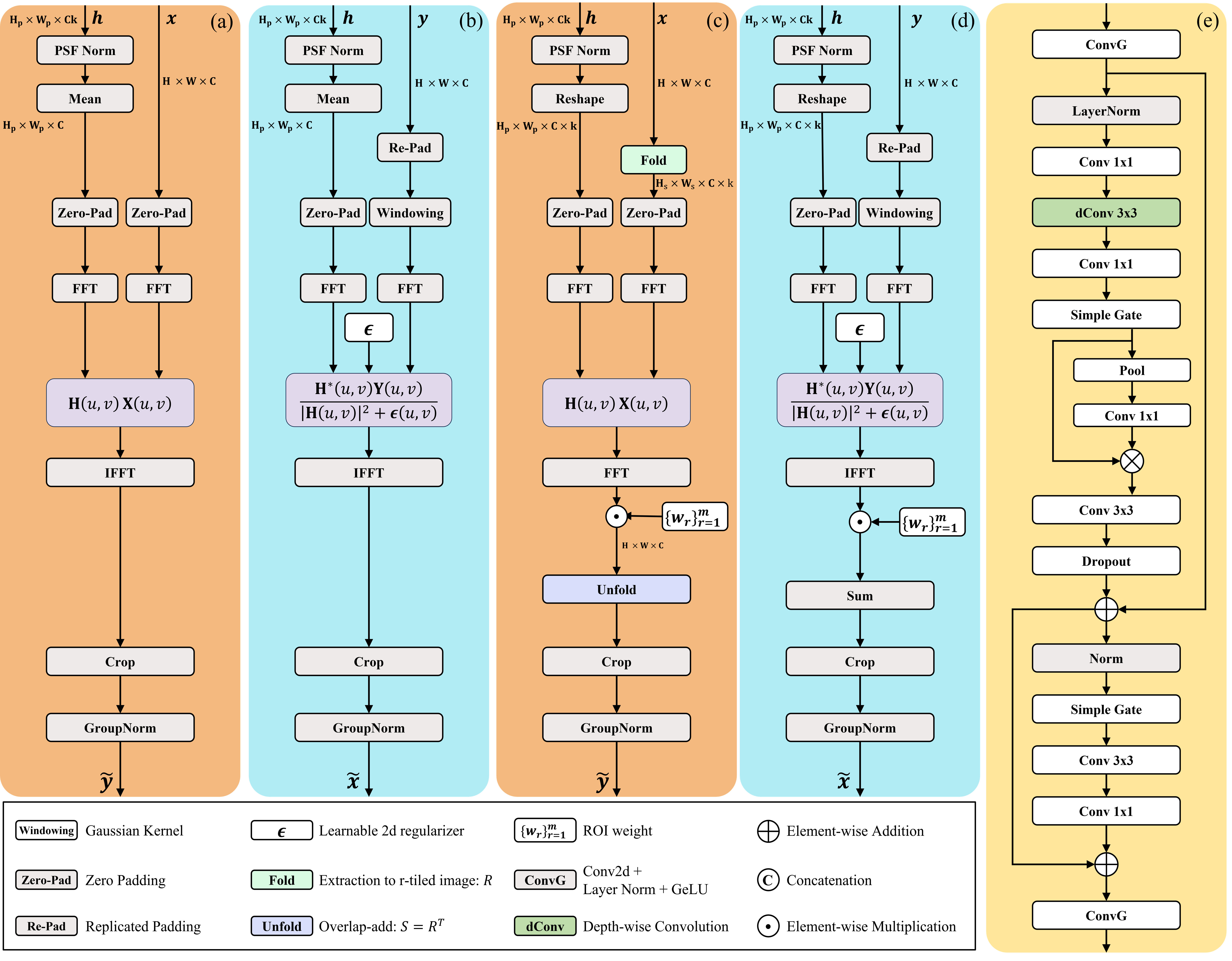}
  \caption{
Detailed flow of system operators and the refinement block used in IFIB.
(a) Basic FSO.
(b) Basic ISO.
(c) SV FSO.
(d) SV ISO.
(e) Refinement block.}
  \label{fig:detail}
\end{figure}

We also considered a fully shift-variant formulation of the forward operator, analogous to ISO, by decomposing $\hat{x}$ into overlapping tiles, padding each tile to the local PSF, convolving locally, and reassembling via normalized overlap-add:
\begin{equation}
\tilde{y}
\;=\;
\sum_{r=1}^{k}
S_r \Big( \big( R_r \hat{x} \big) * h_r \Big),
\label{eq:FSO-sv-P}
\end{equation}
where $R_r$ extracts the $r$-th tile and $S_r \!=\! R_r^\top$ denotes overlap-add with normalization by the local coverage count to avoid seams.
In practice, however, this design increases computational overhead despite efforts to optimize tiling, and a memory-efficient fold/unfold implementation only partially mitigates this. Since the forward operator in IFIN primarily serves to preserve measurement-domain properties rather than to synthesize high-fidelity outputs, providing it in a simplified, shift-invariant (averaged) form reduces model mismatch while retaining the necessary measurement-consistency signal. For these reasons, we adopt the shift-invariant forward operator in our main design. We anticipate that more precise yet simplified variants of the forward operator can be integrated when available, further improving fidelity without incurring significant overhead. 
A detailed flow of the system operators, including the shift-variant FSO, is provided in \cref{fig:detail}.

\section{Expansion of System Variance}
\label{app:expansion}
Depending on the optical configuration, the same PSF field can be re-indexed along an additional axis.

Concretely, we write
\[
h_r[a,b;\, q], \quad r=1,\dots,s^2,\ \ q\in\mathcal{Q},
\]
where $r$ indexes lateral regions (field dependence) and $q$ indexes the non-lateral axis (e.g., depth $z$, wavelength $\lambda$, or time $t$).  
For scenarios where the PSF varies both laterally and axially, we adopt a 5D depth-space-variant PSF:
\[
h_{i,j}[a, b;\, z].
\]
The corresponding forward model is
\begin{equation}
y[i,j] \;=\; 
\sum_{z=1}^{N_z}\;\sum_{a,b}
h_{i,j}[a,b;\,z]\; x[i-a,j-b, z] \;+\; \eta[i,j],
\label{eq:5d-psf}
\end{equation}
where $x[\cdot,\cdot,z]$ denotes the scene slice at depth $z$. For purely field-dependent blur we set $|\mathcal{Q}|{=}1$ and recover the 2D case. This re-indexing keeps the forward/inverse operators unchanged in form while allowing IFIN to adapt the PSF dimension to the underlying system; in this work we focus on the 2D shift-variant setting.

\section{Computational Cost}
Spatially varying operators improve fidelity but reduce parameter sharing, which naturally increases memory usage and runtime relative to the shift-invariant case. To quantify this, we benchmark representative methods with batch size $1$ on a single NVIDIA RTX A6000 GPU, measuring the number of trainable parameters, FLOPs, peak GPU memory, and wall-clock inference time for a single forward pass. The results are summarized in \cref{tab:complexity_all}. Importantly, IFIN ($k{=}1$) already outperforms all baselines in the main quantitative comparison with a parameter count, FLOPs, and memory comparable to strong baselines such as NAFNet, MWDNs, and LensNet, so its accuracy gain cannot be attributed to a substantially larger model. While increasing the number of learnable PSFs $k$ raises parameters and complexity, moderate values remain comparable in FLOPs to other hybrid baselines and provide the best trade-off between quality and cost (see \cref{tab:ablation_psfs} for the corresponding accuracy trends).

The higher memory footprint and slower runtime of IFIN relative to baselines stem mainly from the FFT-based convolutions and deconvolutions used in FSO and ISO. These operators require padding to avoid wrap-around artifacts, and both the padded region and the need to maintain multiple PSF-specific reconstructions in the ISO branch cause the cost to scale with $k$ and the spatial extent of the PSFs. To mitigate this, we crop each PSF to retain only its effective region. This design keeps the overhead manageable while retaining most of the gains from spatially varying modeling.
\begin{table}[tb]
\caption{
Complexity comparison across architectures on DiffuserCam.
We report the number of trainable parameters, FLOPs, peak VRAM, and inference time per image (batch size $1$) on an NVIDIA RTX A6000 GPU.
}
\label{tab:complexity_all}
\begin{center}
\begin{adjustbox}{max width=\linewidth}
\begin{tabular}{lcccc}
\toprule
\bf Method & \bf Params (M) & \bf FLOPs (G)
           & \bf VRAM (GB)
           & \bf Time (ms)\\
\midrule
NAFNet          & 29.2  & 32.4  & 0.257 & 24.65 \\
MWDNs           & 22.1  & 106.4 & 0.422 & 12.81 \\
UPDN            & 3.7   & 24.5  & 0.326 & 60.66 \\
LensNet         & 31.6  & 147.2 & 0.529 & 43.00 \\
MoDL            & 8.5   & 119.1 & 0.202 & 36.60 \\
IFIN ($k{=}1$)  & 18.2  & 55.0  & 0.464 & 142.35 \\
IFIN ($k{=}4$)  & 66.2  & 62.4  & 1.257 & 156.33 \\
IFIN ($k{=}9$)  & 155.9 & 82.4  & 2.614 & 180.37 \\
IFIN ($k{=}16$) & 301.5 & 126.9 & 4.590 & 215.97 \\
\bottomrule
\end{tabular}
\end{adjustbox}
\end{center}
\end{table}

\section{Discussion and Limitations}
Across the three benchmarks (DiffuserCam, WiderCam, and MWNet), IFIN achieves the best scores in all metrics. In terms of PSNR, it improves over the strongest prior learning-based method on each dataset (UPDN on DiffuserCam, MoDL on WiderCam and MWNet) by +1.63\,dB, +0.65\,dB, and +2.58\,dB, respectively. These gains are accompanied by consistently lower LPIPS and higher SSIM, indicating that the improvements reflect both sharper details and better perceptual quality. The advantages are particularly pronounced near the field periphery and on the MWNet benchmark, where large PSFs severely degrade purely CNN-based inversion.

We attribute these improvements to the way the reconstruction network is built around the known forward model and its inverse. Rather than relying on a single inversion stage, IFIN utilizes the integrated forward--inverse operators with learned feature transforms so that measurement-domain consistency is enforced throughout the feature hierarchy, while learnable shift-variant operators driven by a multi-scale PSF field allow the network to adapt to system mismatch.

\subsubsection{Generality beyond a single modality.}
The integrated forward--inverse formulation is not tied to a particular optical prototype or to a specific PSF parameterization. More generally, IFIN only requires that the target imaging system admit a differentiable forward model together with a compatible inverse, pseudo-inverse, or adjoint-like operator that can be embedded into the network hierarchy. In the lensless imaging experiments of the main paper, these operators are instantiated by PSF-based convolution and deconvolution modules. In the inline holography experiment, by contrast, the same architecture is adapted by replacing the PSF-conditioned physical modules with diffraction-based forward and back-propagation operators. This demonstrates that the proposed framework should be viewed as a general forward--inverse integration template rather than as a design limited to calibrated convolutional imaging systems.

This generality, however, still rests on several modeling assumptions. First, the forward operator used during training should provide a sufficiently faithful approximation to the measurement formation process encountered at inference. Second, the corresponding inverse or adjoint operator should be differentiable and numerically stable enough to be inserted as a layer. When these assumptions are significantly violated---for example, under strong nonlinear sensor effects, severe model mismatch, inaccurate system parameters, or noise statistics that differ substantially from those assumed during training---the reconstruction quality may degrade and the learned behavior may deviate from the intended physics. In such cases, IFIN should be regarded as a flexible architectural template whose physics modules must be adapted to the modality of interest.

The inline holography experiment also illustrates an important practical point: the physical modules do not need to preserve the exact same parameterization across modalities. In that benchmark, the PSF-conditioned modules are removed entirely, and the bidirectional interaction is instead driven by a symmetric propagation pair derived from the holographic forward model. This suggests that the essential contribution of IFIN lies in the hierarchical coupling of forward and inverse physics with learned feature transforms, rather than in any single operator instantiation.

Several extensions follow naturally. One direction is to extend the PSF field with depth-, wavelength-, or time-dependent operator channels for volumetric, hyperspectral, or dynamic imaging, as formalized in \cref{app:expansion}. Another direction is to impose low-rank or separable structure on this extended PSF field to reduce the cost of physical operators, or to parameterize the operators by coordinates or deformation fields so that strong variance can be captured with fewer parameters. Finally, co-designing the hardware---for instance, masks or apertures that yield sparser or more localized PSFs---with the proposed reconstruction could further improve fidelity and make the joint system easier to invert.

\subsubsection{Limitations.}
\begin{figure}[tb]
  \centering
  \includegraphics[width=\linewidth]{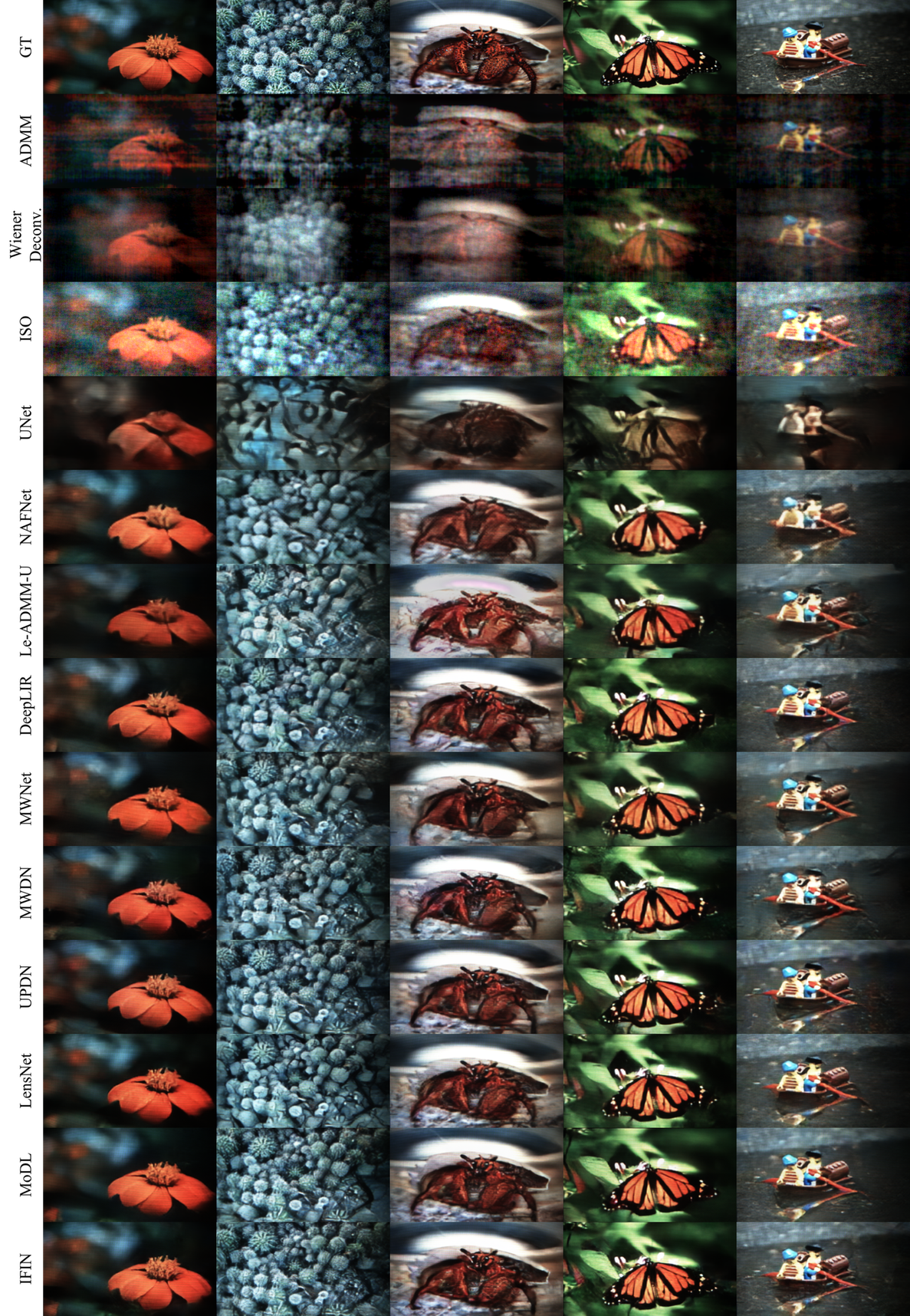}
  \caption{Additional qualitative comparison on the DiffuserCam dataset.}
  \label{fig:appendix_result_DiffuserCam}
\end{figure}

\begin{figure}[tb]
  \centering
  \includegraphics[width=\linewidth]{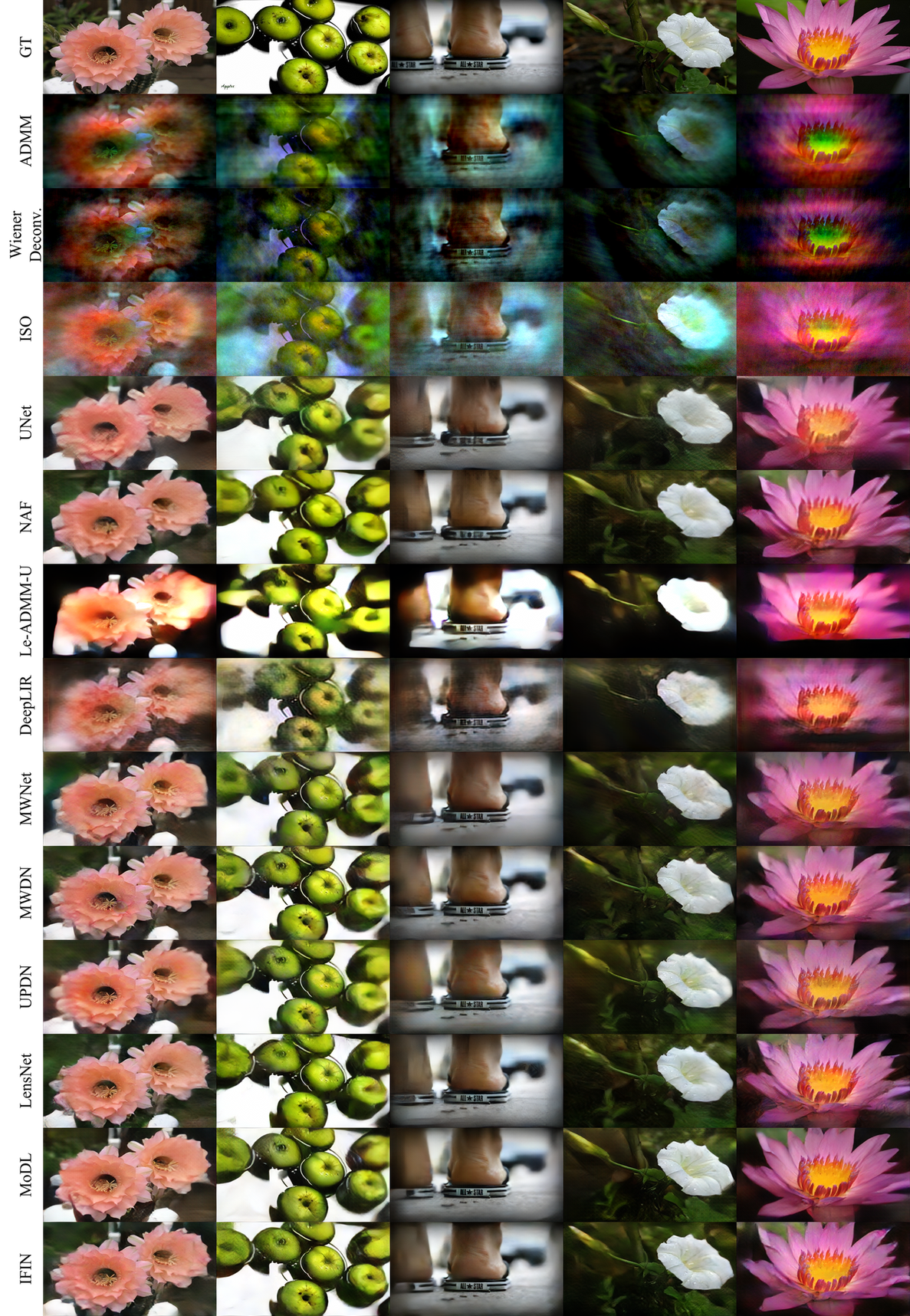}
  \caption{Additional qualitative comparison on the WiderCam dataset.}
  \label{fig:appendix_result_SV}
\end{figure}

While IFIN substantially improves reconstruction quality under severe blur and shift variance, it also has two practical limitations. First, the two-stream, multi-scale forward--inverse architecture with a shift-variant PSF field incurs higher computational and memory cost than simpler feed-forward or unrolled baselines, especially when using a large number of PSFs $k$. As analyzed, the runtime remains acceptable on modern GPUs, but real-time deployment on resource-constrained hardware will require further model compression and operator optimization. Second, our experiments focus on measurements acquired under moderate indoor lighting and a reasonably stable mask--sensor--scene geometry. As in other lensless systems, extreme high-dynamic-range scenes or strong external light can introduce large saturated regions on the sensor, and rapid or large geometry changes can invalidate the assumed forward model and the learned PSF field; in such regimes, information is physically missing and no single-exposure reconstructor, including IFIN and the baselines, can fully recover it. We view reducing the computational footprint and extending robustness to these challenging capture conditions as important directions for future work.
Additional qualitative comparisons on DiffuserCam and WiderCam are shown in \cref{fig:appendix_result_DiffuserCam} and \cref{fig:appendix_result_SV}.

\bibliographystyle{splncs04}
\bibliography{main}